\documentclass[lettersize,journal]{IEEEtran}
\usepackage{amsmath,amsfonts}
\usepackage{algorithmic}
\usepackage{algorithm}
\usepackage{array}
\usepackage[caption=false,font=normalsize,labelfont=sf,textfont=sf]{subfig}
\usepackage{textcomp}
\usepackage{stfloats}
\usepackage{url}
\usepackage{verbatim}
\usepackage{graphicx}
\usepackage{cite}
\usepackage{tikz}
\usetikzlibrary{spy}
\usepackage[pagebackref,breaklinks,colorlinks]{hyperref}
\usepackage[capitalize]{cleveref}
\usepackage{xspace}
\usepackage{multirow}
\usepackage{pifont}

\newcommand{\ie}{{\emph{i.e.}}\xspace}
\newcommand{\vs}{{\emph{vs.}}\xspace}
\newcommand{\eg}{{\emph{e.g.}}\xspace}
\newcommand{\etal}{{\emph{et al.}}\xspace}

\newcommand{\subcref}[1]{{\color{red}#1}}

\newcommand{\myname}{E-DEI\xspace}
\newcommand{\mydata}{PIED\xspace}

\newcommand{\revision}[1]{\textcolor{black}{#1}}

\begin{document}

% \title{Event-based Low-Light Imaging with Dual-Exposure Image Pairs}
\title{Dual-Exposure Imaging with Events}

% \author{Anonymous submission
% \thanks{}
% }

\author{Mingyuan Lin, Hongyi Liu, Chu He, Wen Yang, Gui-Song Xia, and Lei Yu
        % <-this % stops a space
\thanks{M. Lin, H. Liu, C. He, and W. Yang are with the School of Electronic Information, Wuhan University, Wuhan 430072, China. E-mail: \{linmingyuan, mcclhy, chuhe, yangwen\}@whu.edu.cn.}
\thanks{G.-S. Xia and L. Yu are with the School of Artificial Intelligence, Wuhan University, Wuhan 430072, China. E-mail: \{guisong.xia, ly.wd\}@whu.edu.cn.}
\thanks{The research was partially supported by the National Natural Science Foundation of China under Grant 62271354 and the National Cybersecurity Talent and Innovation Base, Wuhan.}
\thanks{Corresponding authors: Chu He and Lei Yu.}
}

\markboth{Journal of \LaTeX\ Class Files}%
{Shell \MakeLowercase{\textit{et al.}}: A Sample Article Using IEEEtran.cls for IEEE Journals}


\maketitle

\begin{abstract}
% By combining complementary benefits of short- and long-exposure images, Dual-Exposure Imaging (DEI) enhances the image quality in low-light scenarios. However, existing DEI approaches inevitably suffer from producing artifacts due to the spatial displacement from scene motion and image feature discrepancies from different exposure times. To tackle this problem, we propose a novel Event-based DEI (\myname) algorithm, which reconstructs high-quality images from dual-exposure image pairs and events, leveraging the high temporal resolution of event cameras to provide accurate inter-/intra-frame dynamic information. Specifically, we decompose this complex task into an integration of two sub-tasks, \ie, event-based motion deblurring and low-light image enhancement tasks, which guides us to design \myname network as a dual-path parallel feature propagation architecture. We propose a Dual-path Feature Alignment and Fusion (DFAF) module to effectively align and fuse features extracted from dual-exposure images with the assistance of events. Furthermore, we build a new real-world Dataset containing Paired low-/normal-light Images and Events (\mydata). Experiments on synthetic and real-world datasets demonstrate the superiority of our method. The code and dataset are available at \url{https://github.com/mingyuan-lin/E-DEI/}.

By combining complementary benefits of short- and long-exposure images, Dual-Exposure Imaging (DEI) enhances image quality in low-light scenarios. However, existing DEI approaches inevitably suffer from producing artifacts due to spatial displacement from scene motion and image feature discrepancies from different exposure times. To tackle this problem, we propose a novel Event-based DEI (\myname) algorithm, which reconstructs high-quality images from dual-exposure image pairs and events, leveraging high temporal resolution of event cameras to provide accurate inter-/intra-frame dynamic information. Specifically, we decompose this complex task into an integration of two sub-tasks, \ie, event-based motion deblurring and low-light image enhancement tasks, which guides us to design \myname network as a dual-path parallel feature propagation architecture. We propose a Dual-path Feature Alignment and Fusion (DFAF) module to effectively align and fuse features extracted from dual-exposure images with assistance of events. Furthermore, we build a real-world Dataset containing Paired low-/normal-light Images and Events (\mydata). Experiments on multiple datasets show the superiority of our method. The code and dataset are available at \url{https://github.com/mingyuan-lin/E-DEI/}.

\end{abstract}

\begin{IEEEkeywords}
Low light, dual-exposure imaging, event camera.
\end{IEEEkeywords}

\section{Introduction}

\begin{figure}[t]
    \centering
    \includegraphics[width=\linewidth]{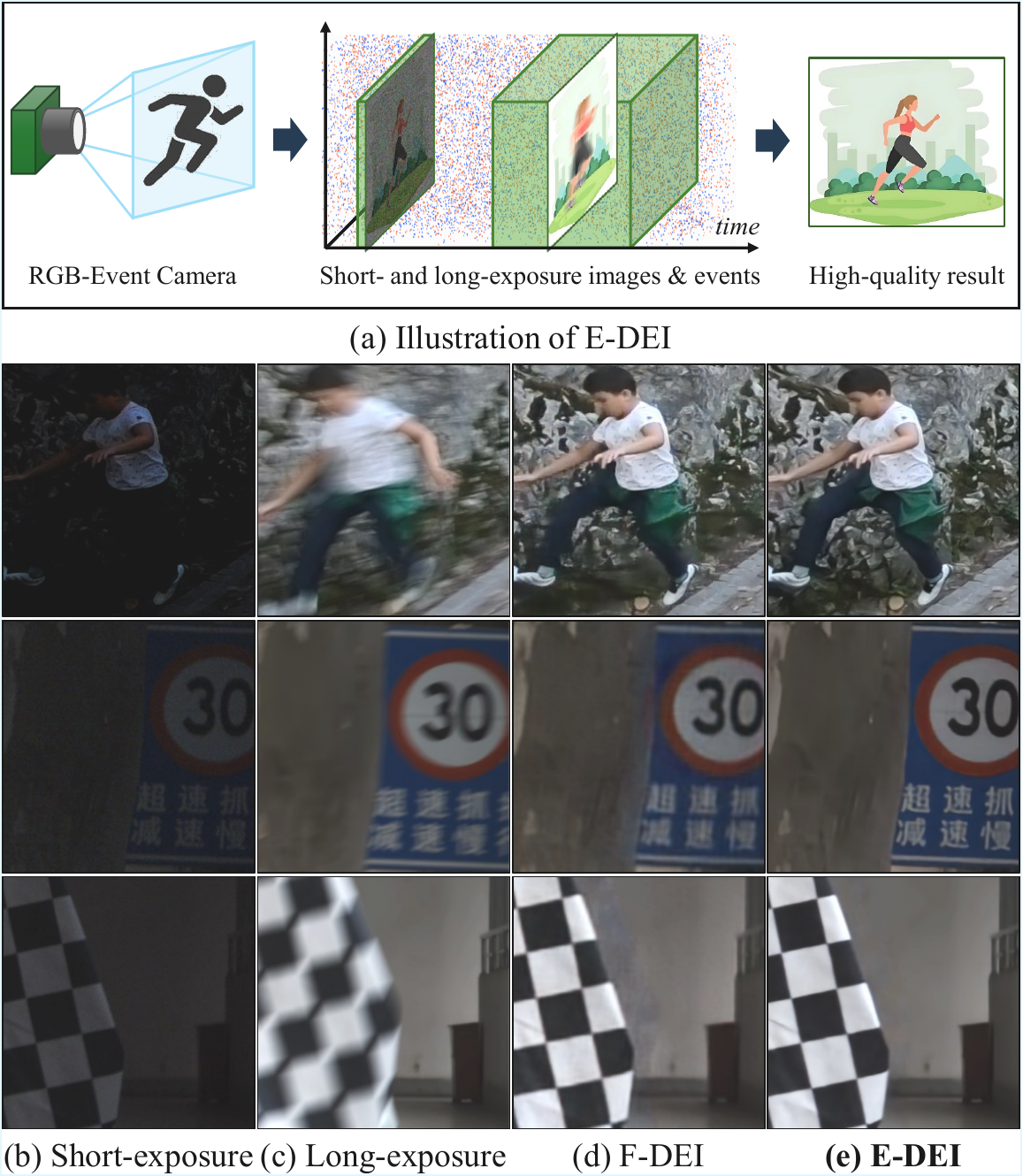}
    \caption{(a) Illustration of the proposed Event-based Dual-Exposure Imaging (\myname) strategy. (b-c) The short- and long-exposure image pairs are continuously captured under the dynamic scenes. The corresponding image reconstruction results via (d) the state-of-the-art Frame-based DEI (F-DEI) method and (e) our \myname. Due to the missing inter/intra-frame motion information, F-DEI suffers from texture blur and artifacts, while \myname can successfully generate the high-quality visual images.}
    \label{fig:first}
\end{figure}

\IEEEPARstart{I}{n} the field of computational photography, imaging under harsh conditions, \ie, highly dynamic scenes involving either camera or object motion, remains a formidable challenge \cite{li2021low,wang2021seeing,zhang2025exposure}. Recent studies have introduced the Dual-Exposure Imaging (DEI) strategy, exploring the simultaneous utilization of sequentially captured short- and long-exposure image pairs for high-quality imaging, especially under the low-light environments \cite{yuan2007image,mustaniemi2020lsd2,chang2021low,zhao2022d2hnet}. Compared with the conventional Single-Exposure Imaging (SEI) strategy that struggles to balance Signal-to-Noise Ratio (SNR), texture sharpness, and color fidelity \cite{kim2020proactive,tomasi2021learned,zhang2022exploring,han2023camera,shin2022drl}, DEI attempts to combine the respective advantages of the image pairs to enable the reconstructed images with both high SNR and sharp textures. However, as shown in \cref{fig:first}, two fundamental challenges hinder practical implementation:
\begin{itemize}
    \item \textbf{Missing dynamic information.} In highly dynamic scenes, the missing inter- and intra-frame motion information due to the sequential image acquisition at fixed frame rates of standard cameras causes the spatial displacement between dual-exposure image pairs \cite{shen2021spatial}.
    \item \textbf{Inherent discrepancies between images.} The varying exposure times of the image pairs collected through DEI lead to multiple inherent discrepancies in brightness, noise level, and texture sharpness, which significantly degrade the image reconstruction quality \cite{zhang2024ntire}.
\end{itemize}

% \begin{figure}[t]
%     \centering
%     \includegraphics[width=\linewidth]{figs/pipeline/fist.png}
%     \caption{Illustrations of the improvement from Single-Exposure Imaging (SEI), \ie, (a) low-light image enhancement for short-exposure images and (b) motion deblurring for long-exposure images, and (c) conventional Dual-Exposure Imaging (DEI) tasks to (d) our proposed Event-based Dual-Exposure Imaging (EDEI). Although recent DEI methods attempt to improve the low-light imaging quality by simultaneously utilizing consecutively captured short- and long-exposure images, they fail to handle the spatial misalignment caused by scene motion. In this paper, we propose to further promote the performance of DEI with the assistance of motion information recorded by event cameras.}
%     \label{fig:first}
% \end{figure}

\IEEEpubidadjcol

In this paper, we propose to utilize the event camera to enhance the performance of DEI. Event cameras are neuromorphic sensors that encode the scene brightness variations and asynchronously report event data with extremely high temporal resolution and extremely low latency (in the order of $\mu s$) \cite{lichtsteiner128Times1282008,gallego2020event,wang2024asynchronous}. They are free of the notion of exposure time and thus do not suffer from the inevitable trade-off between the motion blur in long-exposure images and low SNR in short-exposure images \cite{liang2023coherent}. Instead of only using intensity images, the result in \cref{fig:first}\subcref{(e)} demonstrates that the temporal consistency and low latency of the event data benefits in bridging the gap between dual-exposure images and meanwhile supplementing the missing inter- and intra-frame motion information, unlocking the potential to reconstruct the sharp and clean images from the noisy and blurry observations. 

Specifically, we propose a novel network of Event-based Dual-Exposure Imaging (\myname). It is designed as a dual-path feature propagation and supervision architecture, inspired by the observation that high-quality imaging can be achieved by: a) \textbf{motion deblurring} from the long-exposure image and corresponding events, where the short-exposure image serves as the temporal reference, and b) \textbf{low-light image enhancement} by fusing short- and long-exposure image pairs, where the aligned long-exposure image serves as the intensity reference and events as the texture reference. Accordingly, we propose a Dual-path Feature Alignment and Fusion (DFAF) module. In the deblurring path, event streams are utilized to align the long-exposure image to the short-exposure reference without the direct short-long feature information fusion. Subsequently, the enhancement path integrates the aligned long-exposure image with the short-exposure one for the high-quality restoration. Through employing simultaneous supervisions of dual paths, our \myname can explicitly learn the event-based motion deblurring of the long-exposure image while promoting complementary fusion and enhancement of features of events and dual-exposure images.

Although several low-light datasets containing real-world events have emerged recently, they are not well suited for the DEI task, \eg, the RLED dataset \cite{liu2024seeing} lacks real-world low-light images, while the EVFI-LL dataset \cite{zhang2024sim} does not provide sharp and noise-free ground-truth images. Although the SDE dataset \cite{liang2024towards} collects temporally and spatially aligned low- and normal-light sharp image pairs using an ND8 filter and a robotic arm, limitations remain since it only records camera ego-motion and lacks high-quality data as it is acquired with a DAVIS346 sensor. For effective training, we build a hybrid camera system where two RGB cameras and one event camera are co-optical axis aligned with two 50:50 beam splitters. Thus, a new real-world Dataset consisting of Paired normal- and low-light sharp Images and corresponding Events (\mydata) with high temporal and spatial resolutions is constructed to facilitate the future research.

The main contributions can be summarized as:
\begin{itemize}
    \item We propose a novel network of Event-based Dual-Exposure Imaging (\myname), aiming to reconstruct high-quality images from sequentially captured long- and short-exposure image pairs with the assistance of low-latency event data.
    \item We discover two keys to achieve high-quality imaging are motion deblurring using long-exposure images and low-light image enhancement using short-exposure images, and design a Dual-path Feature Alignment and Fusion (DFAF) module to facilitate fusion and enhancement of features of events and dual-exposure images.
    \item We build a new real-world Dataset composed of Paired normal- and low-light sharp Images and corresponding Event streams (\mydata), containing various scenes and diverse motion types. Extensive qualitative and quantitative comparisons on multiple datasets demonstrate the effectiveness and robustness of our proposed \myname. \revision{The code and dataset are available at \url{https://github.com/mingyuan-lin/E-DEI/}.}
\end{itemize}

\section{Related Work}
\subsection{Single-Exposure Image Restoration}
Various low-light image reconstruction strategies have been developed to achieve high-quality results. They can be roughly categorized into two classes based on the exposure time of input images: image enhancement for short-exposure inputs \cite{wei2018deep,wu2022uretinex,fu2023learning,xu2022snr,xu2023low} and motion deblurring for long-exposure ones \cite{zhang2021exposure,fang2023self,cho2021rethinking,zhang2023multi,zamir2021multi,chen2021hinet,kupyn2018deblurgan,ren2023multiscale}. For short-exposure image enhancement, retinex-based methods decompose low-light images into reflectance and illumination components, then brighten up the illumination \cite{wei2018deep,wu2022uretinex,fu2023learning}. End-to-end approaches learn brute-force mapping functions from degraded images to normal-light counterparts, taking advantage of a large amount of data \cite{xu2022snr,xu2023low}. However, these methods often suffer from the color distortion due to the absence of the normal-light intensity reference. Although extending exposure times helps to provide higher SNR and more reliable intensity information, texture erasure, as another type of distortion, becomes dominant in dynamic scenes. One of the most popular deblurring methods is to employ neural networks and adopt multiple techniques, including modeling the blurry image as convolutions of the latent image and blur kernels \cite{zhang2021exposure,fang2023self}, coarse-to-fine strategies \cite{cho2021rethinking,zhang2023multi}, multi-stage architectures \cite{zamir2021multi,chen2021hinet}, and generative methods \cite{kupyn2018deblurgan,ren2023multiscale}. However, they still struggle with the intra-frame motion ambiguity, particularly under non-uniform motion in real-world scenarios.

\subsection{Dual-Exposure Image Restoration}
Several methods exploit the complementarity of short- and long-exposure image pairs for the better restoration \cite{yuan2007image,gu2020blur,shen2021spatial,zhang2022self,shekarforoush2023dual,rim2024deep,mustaniemi2020lsd2,chang2021low,zhao2022d2hnet}. Some works rely on the strict assumption that the dual-exposure images are spatially aligned, which is not always fulfilled in real-world dynamic scenes, \eg, camera shaking and moving objects \cite{yuan2007image,zhang2022self,shekarforoush2023dual,rim2024deep}. Mustaniemi \etal \cite{mustaniemi2020lsd2} adopt a lightweight architecture, directly feeding the concatenated dual-exposure images into a U-Net style network for reconstruction. To address the spatial displacement caused by the missing inter-frame motion information due to the sequential image acquisition at fixed frame rates of standard cameras, Shen \etal \cite{shen2021spatial} attempt to explicitly estimate the inter-frame optical flow to aggregate short- and long-exposure images, while Chang \etal \cite{chang2021low} and Zhao \etal \cite{zhao2022d2hnet} employ the deformable convolution \cite{zhu2019deformable} as the implicit alignment strategy. However, the lack of inter-frame motion information in dynamic scenes and the complex discrepancies in brightness, noise distribution hinder precise spatiotemporal alignment across the dual exposures, leading to residual ghosting artifacts and detail distortions in reconstructed images.

\subsection{Event-based Image Enhancement}
% Differing from the standard cameras that capture intensity images at fixed frame rates, event cameras respond to the brightness change and emit asynchronous event streams \cite{gallego2020event}. This mechanism leads to many promising properties, \eg, low latency and high dynamic range (HDR), which have been utilized in various Event-based Single-Exposure Imaging (E-SEI) tasks. For instance, the recorded intra- and inter-frame motion information benefits motion deblurring \cite{sun2022event,zhang2022unifying,zhang2023generalizing,yang2024motion}, video frame interpolation \cite{tulyakov2021time,he2022timereplayer,ding2024video}, and rolling shutter correction \cite{zhou2022evunroll,erbach2023evshutter,lin2025self} tasks. Recent works attempt to leverage the HDR advantages of event cameras for the low-light image enhancement \cite{jiang2023event,liang2023coherent,liang2024towards,zuo2024cross} task. However, E-SEI methods remain constrained by inherent limitations, \eg, the color fidelity degradation and the irreconcilable trade-off between the texture sharpness in long-exposure images and high SNR in short-exposure ones, significantly hindering real-world applicability. Therefore, leveraging the temporal consistency of event cameras, which is afforded by their intrinsic exposure-free property, to advance the \myname task and further boost image quality is one of the key aspects of this paper.

Differing from the standard cameras that capture intensity images at fixed frame rates, event cameras respond to the brightness change and emit asynchronous event streams \cite{gallego2020event}. \revision{This mechanism leads to many promising properties and benefits several tasks: 1) High temporal resolution. Events can record the intra-/inter-frame continuous motion and texture information, which is crucial for various single-exposure imaging tasks, \eg, motion deblurring \cite{sun2022event,zhang2022unifying,zhang2023generalizing,yang2024motion}, video super-resolution \cite{xiao2024asymmetric,xiao2024event,xiao2025event}, frame interpolation \cite{tulyakov2021time,he2022timereplayer,ding2024video}, and rolling shutter correction \cite{zhou2022evunroll,erbach2023evshutter,lin2025self}. 2) Exposure-independent and temporal-consistent statistics. The event generation does not suffer from the inevitable trade-off between the motion blur and low SNR, thus can build the temporal correspondence among multi-exposure images. Recent works attempt to leverage this advantage for the High Dynamic Range (HDR) imaging task \cite{shaw2022hdr,messikommer2022multi,samra2023high,guo2024event}. However, existing event-based methods do not consider the coupled degradations of cross-exposure image differences and the spatial misalignment in dynamic scenes, significantly hindering real-world applicability. Therefore, leveraging the high temporal resolution and the exposure-independent consistency of event cameras in the \myname task to further boost imaging quality is one of the key aspects of this paper.}

\section{Problem Formulation}
\subsection{Intensity Image Formation}
\revision{As illustrated in \cite{chen2018reblur2deblur}, the process of traditional cameras capturing long-exposure blurry images can be formulated as the average of the instant images within the exposure period $\mathcal{T}=[t_b,t_e]$. A general model of image formation is given by, 
\begin{equation}
    I_l(\mathcal{T})=\frac{1}{T}\int_{t\in\mathcal{T}}L(t)dt,
    \label{eq:i_l}
\end{equation}
where $I_l$ means the long-exposure image with exposure time $T=t_e-t_b$ and $L(t)$ denotes the latent instant intensity at time $t \in \mathcal{T}$. Although imaging with long exposure time can achieve the high SNR and color fidelity, the significant motion blur inevitably occurs, making it difficult to ensure sharp and detailed textures in the highly dynamic scenes \cite{kim2022mssnet,zhang2021exposure}.}

On the other hand, shortening the exposure time helps to alleviate blur. However, in particular, the short-exposure images are degraded by the insufficient photosensitive period and noise \cite{healey1994radiometric,izadi2023image}. Current studies \cite{lv2021attention,zhang2021learning} synthesize the low-light image darkening by the combination of gamma correction and linear scaling and adopting the signal-dependent Gaussian distribution to synthesize the noise as,
\begin{equation}
    \begin{aligned}
        I_s(t_s)&\sim\mathcal{N}(J(t_s), \sigma_pJ(t_s)+\sigma^2_g), \operatorname{with} \\
        J(t_s)&=\beta\times(\alpha\times L(t_s))^\gamma,
    \end{aligned}
    \label{eq:i_s}
\end{equation}
where $I_s$ is the short-exposure instant image at time $t_s$, with the exposure length short enough to be ignored. $\mathcal{N}$ means the Gaussian distribution, and $\alpha,\beta,\gamma$ are the darkening parameters and $\sigma_p,\sigma_g$ are the noise parameters. 

\begin{figure}
    \centering
    \begin{tabular}{c}
        \hspace{-3mm} \includegraphics[width=\linewidth]{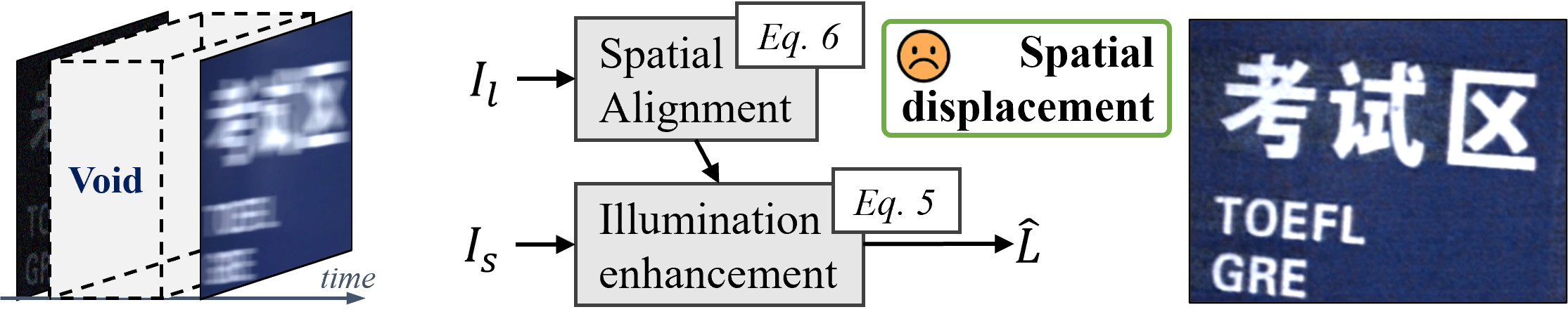} \\
        {\small (a)} \\
        \hspace{-3mm} \includegraphics[width=\linewidth]{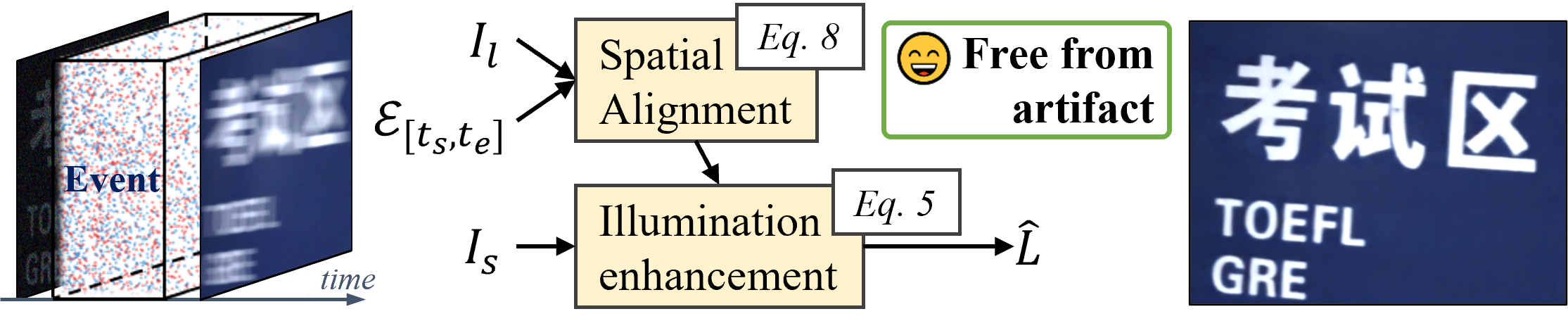} \\
        {\small (b)} \\
    \end{tabular}
    \caption{Comparison between (a) existing Frame-based Dual-Exposure Imaging (F-DEI) and (b) our Event-based Dual-Exposure Imaging (\myname). The key idea of our \myname is to introduce the high temporal resolution event data to complement the missing intra- and inter-frame motion information. As a result, our algorithm produces sharp and clear images without artifacts.}
    \label{fig:dei}
\end{figure}

\subsection{Frame-based Dual-Exposure Imaging}
By taking the complementary advantages of short- and long-exposure image pairs, recent methods \cite{gu2020blur,chang2021low,zhao2022d2hnet} introduce the Frame-based Dual-Exposure Imaging (F-DEI) strategy for image restoration with high SNR, color fidelity, and sharp textures, defined by, 
\begin{equation}
    \hat{L}(t_s) = \operatorname{F-DEI}(I_{s}(t_s), I_{l}(\mathcal{T})),
    \label{eq:fdei}
\end{equation}
where $\hat{L}$ denotes the predicted image result at the same time as $I_s$. \revision{Assuming we are shooting the static scenes, the long- and short-exposure images $I_l, I_s$ are spatially aligned at the pixel level. Thus, the F-DEI task can be simply realized via the ratio between $I_{l}$ and $I_{s}$ as,
\begin{equation}
    \hat{L}(t_s) = \frac{I_{l}(\mathcal{T})}{I_{s}(t_s)} \cdot I_{s}(t_s).
    \label{eq:fdei_s}
\end{equation}}

However, motion in the real-world scenes breaks the above assumption, \ie, $I_l$ and $I_s$ may not share the same reflection $R$. Thus, as shown in \cref{fig:dei}\subcref{(a)}, the key point of the F-DEI task is to estimate the pixel-level spatial mapping from $I_l$ to $I_s$, and \cref{eq:fdei_s} can be rewritten as,
\begin{align}
    \hat{L}(t_s) &= \frac{\hat{L}_{l}(t_s)}{I_{s}(t_s)} \cdot I_{s}(t_s), \operatorname{with} \label{eq:fdei_d1} \\
    \hat{L}_{l}(t_s) &= \mathcal{M}_{l \rightarrow s}(I_{l}(\mathcal{T})),  \label{eq:fdei_d2}
\end{align}
where $\mathcal{M}_{l \rightarrow s}(\cdot)$ is the spatial mapping function from the domain of $I_l$ to that of $I_s$, and $\hat{L}_{l}$ is the predicted image from the long-exposure image. This is extremely challenging due to the coupled degradation of missing inter-/intra-frame motion information and inherent discrepancies between $I_l$ and $I_s$. 

\subsection{Event-based Dual-Exposure Imaging}
Unlike conventional cameras, each pixel of an event camera responds to changes in the brightness and generates streams of asynchronous events \cite{lichtsteiner128Times1282008,gallego2020event,wang2024asynchronous}. For an event camera, the event $e_i=(t_i,\mathbf{x}_i,p_i)$ is triggered whenever the logarithmic pixel intensity difference reaches a threshold $C$, \ie,
\begin{equation}
    \log(L(t_i,\mathbf{x}_i))-\log(L(f,\mathbf{x}_i))=p_i \cdot C,
    \label{eq:ll_pc}
\end{equation}
where $L(t,\mathbf{x})$ denotes the latent intensity at time $t$ at the pixel position $\mathbf{x}$ and $p\in\{+1,-1\}$ denotes the polarity showing the direction of brightness change. \revision{The high temporal resolution (in the order of $\mu s$) of event cameras assists to supplement lost inter- and intra-frame scene motion information, which is crucial for solving the spatial mapping problem in DEI. Thus, as shown in \cref{fig:dei}\subcref{(b)}, events can provide the temporal clue to help solve the spatial mapping problem, rewriting \cref{eq:fdei_d2} as,
% \begin{equation}
%     \begin{aligned}
%         \hat{L}_{l;\Omega_s}(t_s) &= \mathcal{M}_{\Omega_l\rightarrow\Omega_s}(I_{l;\Omega_l}(\mathcal{T}), \mathcal{E}_{[t_s, t_e]}), \operatorname{with} \\
%         \mathcal{E}_{[t_s, t_e]} &= \{e(\tau)|\tau\in[t_s, t_e]\} \\
%     \end{aligned}
%     \label{eq:edei_d1}
% \end{equation}
\begin{equation}
    \hat{L}_{l}(t_s) = \mathcal{M}_{l \rightarrow s}(I_{l}(\mathcal{T}), \mathcal{E}_{[t_s, t_e]}), 
    \label{eq:edei_d1}
\end{equation}
% \begin{align}
%     \hat{L}_{l}(t_s) &= \mathcal{M}_{l \rightarrow s}(I_{l}(\mathcal{T}), \mathcal{E}_{[t_s, t_e]}), \label{eq:edei_d1} \\
%     \hat{L}(t_s) &= \frac{\hat{L}_{l}(t_s)}{I_{s}(t_s)} \cdot I_{s}(t_s), \label{eq:edei_d2}
% \end{align}
% \begin{equation}
%     \begin{aligned}
%         \hat{L}_{\Omega_s}(t_s) &= \frac{\mathcal{M}_{\Omega_l\rightarrow\Omega_s}(I_{l;\Omega_l}(\mathcal{T}), \mathcal{E}_{[t_s, t_e]})}{I_{s;\Omega_s}(t_s)} \cdot I_{s;\Omega_s}(t_s), \operatorname{with} \\
%         \mathcal{E}_{[t_s, t_e]} &= \{e(\tau)|\tau\in[t_s, t_e]\},
%     \end{aligned}
% \end{equation}
where $\mathcal{E}_{[t_s, t_e]} = \{e(\tau)|\tau\in[t_s, t_e]\}$. $t_s$ indicates the timestamp of $I_{l}$, $\hat{L}_{l}$, and $\hat{L}$, while $t_e$ is the ending time of the exposure time of $I_{l}$, assuming that the acquisition of the short-exposure image precedes that of the long-exposure one ($t_s<t_b<t_e$).}

% In this paper, we propose a novel task of Event-based Dual-Exposure Imaging (\myname), which receives the short- and long-exposure image pairs $I_{s;\Omega_s}(t_s), I_{l;\Omega_l}(\mathcal{T})$ and the corresponding event streams $\mathcal{E}$ as inputs, and outputs a latent high-quality image $\hat{L}_{\Omega_s}(t_s)$ as,
% \begin{equation}
%     \hat{L}_{\Omega_s}(t_s)=\operatorname{\myname}(I_{s;\Omega_s}(t_s), I_{l;\Omega_l}(\mathcal{T}), \mathcal{E}_{[t_s, t_e]}),
% \end{equation}

% Combing \cref{eq:fdei_d1,eq:edei_d1,eq:edei_d2}, we propose a new task of Event-based Dual-Exposure Imaging (\myname) as ($t_s$ and $\mathcal{T}$ are omitted for readability),
% \begin{equation}
%     \hat{L}=\operatorname{\myname}(I_s, I_l, \mathcal{E}_{[t_s, t_e]}), 
%     \label{eq:edei}
% \end{equation}
% which receives the dual-exposure image pairs $I_s, I_l$ and the corresponding event streams $\mathcal{E}$ as inputs, and outputs a latent high-quality image $\hat{L}$. Specifically, we design a novel network of \myname in \cref{sec:method} to realize the \cref{eq:edei}. 

\revision{Through \cref{eq:fdei_d1,eq:edei_d1}, we propose a novel task of Event-based Dual-Exposure Imaging (\myname) as,}
\begin{equation}
    \hat{L}(t_s)=\operatorname{\myname}(I_s(t_s), I_l(\mathcal{T}), \mathcal{E}_{[t_s, t_e]}),
    \label{eq:edei}
\end{equation}
which receives the dual-exposure image pairs $I_s, I_l$ and the corresponding event streams $\mathcal{E}$ as inputs, and outputs a latent high-quality image $\hat{L}$. It motivates us to focus on exploiting an effective network architecture to precisely bridge the spatial gap between the long- and short-exposure images via the event streams, effectively fusing the dual-exposure image and event features for high-quality and fidelity imaging.

% which receives the short- and long-exposure image pairs $I_{s;\Omega_s}(t_s), I_{l;\Omega_l}(\mathcal{T})$ and the corresponding event streams $\mathcal{E}$ as inputs, and outputs a latent high-quality image $\hat{L}_{\Omega_s}(t_s)$ as,
% \begin{equation}
%     \hat{L}_{\Omega_s}(t_s)=\operatorname{\myname}(I_{s;\Omega_s}(t_s), I_{l;\Omega_l}(\mathcal{T}), \mathcal{E}_{[t_s, t_e]}),
% \end{equation}

% However, motion in the real-world scenes breaks the above ideal assumption. Hence, $I_l$ and $I_s$ do not share the same reflection $R$. It causes coupled temporal and spatial degradations:  
% \begin{itemize}
%     \item Extending the exposure time erases the original textures in $I_l$.
%     \item Missing inter-frame motion information leads to the spatial displacement between dual-exposure images.
% \end{itemize}

\section{Method}
\label{sec:method}

\subsection{\revision{Task Analysis and Decomposition}}
\label{sec:task_decomposition}
% In this paper, we propose a novel task of Event-based Dual-Exposure Imaging (\myname), which receives the short- and long-exposure image pairs $I_s, I_l$ and the corresponding events $\mathcal{E}$ as inputs, and outputs a latent high-quality image $\hat{L}$ as,
% \begin{equation}
%     \hat{L}=\operatorname{\myname}(I_s, I_l, \mathcal{E}_{[t_s, t_e]}),
% \end{equation}
% where we adopt a default assumption that the acquisition of the short-exposure image precedes that of the long-exposure one. $t_s$ indicates the timestamp of $I_s$ and $\hat{L}$, while $t_e$ is the ending time of the exposure time of $I_l$. 

We propose to decompose this complex \myname task into an integration of two sub-tasks, \ie, the Event-based Motion Deblurring (E-MD) task and the Event-based Low-light Image Enhancement (E-LIE) task, which guides us to design a dual-path parallel network architecture.

% \begin{figure}
%     \centering
%     \begin{tabular}{c}
%         \hspace{-3mm} \includegraphics[width=\linewidth]{figs/rebuttal/pipeline_overall.png} \\
%         (a) Overview of \myname \\
%         \hspace{-3mm} \includegraphics[width=0.915\linewidth]{figs/rebuttal/pipeline_dfaf.png} \\
%         (b) Details of DFAF module \\
%     \end{tabular}
%     \caption{(a) The overall architecture of the proposed network of Event-based Dual-Exposure Imaging (\myname), which is a dual-path feature propagation and supervision framework. (b) The proposed Dual-path Feature Alignment and Fusion (DFAF) module, where features of the \textit{Deblurring} path are spatially aligned to the short-exposure target time with features of the \textit{Enhancement} path as the temporal reference, and then we integrate features of the two paths for the high-quality reconstruction.}
%     \label{fig:pipeline}
% \end{figure}

\begin{figure*}[t]
    \centering
    \begin{tabular}{c c}
        \hspace{-3mm}\includegraphics[width=0.55\linewidth]{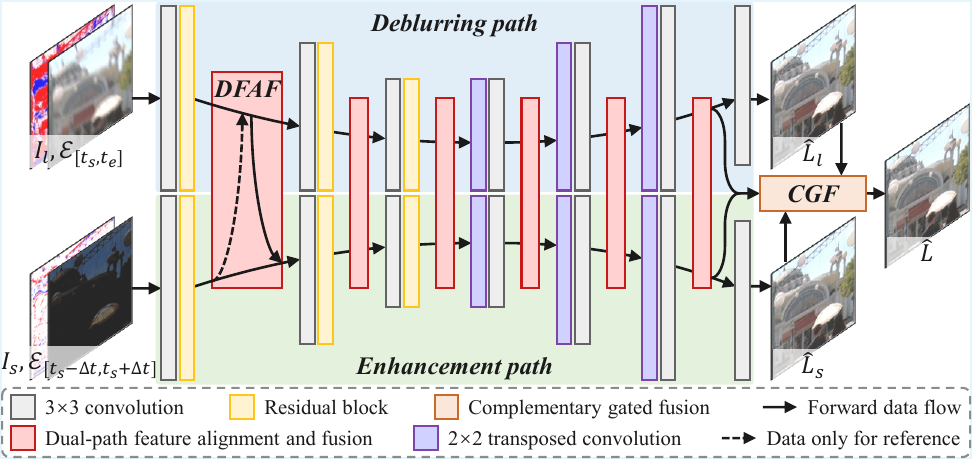}\hspace{-3mm} &  \includegraphics[width=0.42\linewidth]{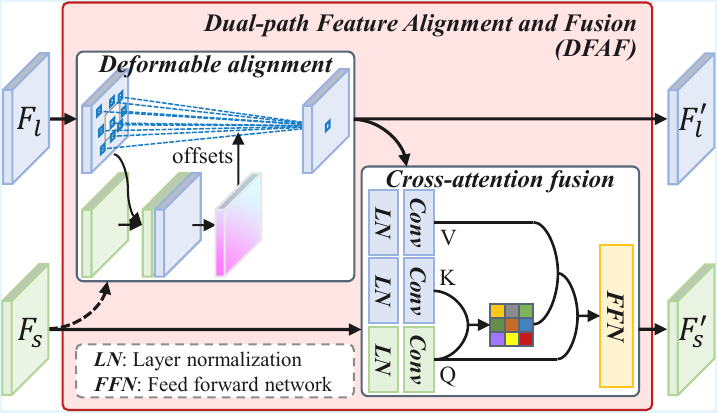} \\
        \revision{\small (a) Overview of \myname} & \revision{\small (b) Details of DFAF module} \\
    \end{tabular}
    \caption{\revision{(a) The overall architecture of the proposed network of Event-based Dual-Exposure Imaging (\myname), which is a dual-path feature propagation and supervision framework. (b) The proposed Dual-path Feature Alignment and Fusion (DFAF) module, where features of the \textit{Deblurring} path are spatially aligned to the short-exposure target time with features of the \textit{Enhancement} path as the temporal reference, and then we integrate features of the two paths for the high-quality reconstruction.}}
    \label{fig:pipeline}
\end{figure*}

\subsubsection{Event-based Motion Deblurring}
\revision{Events can build the temporal connection between the long-exposure blurry image and the short-exposure one.} Hence, with the assistance of events $\mathcal{E}_{[t_s,t_e]}$, we can design the \textit{Deblurring} network to realize the \cref{eq:edei_d1}, aiming to map the long-exposure image $I_l$ to the sharp and clean target $\hat{L}$ at time $t_s$,
\begin{equation}
    \hat{L}_l=\operatorname{E-MD}(I_l, \mathcal{E}_{[t_s,t_e]}| I_s),
    \label{eq:emd}
\end{equation}
where $\hat{L}_l$ indicates the restoration results of $\operatorname{E-MD}(\cdot)$, the event-based motion deblurring function. Different from the existing E-MD methods \cite{zhang2022unifying,yang2024motion}, we introduce the short-exposure image $I_s$, which shares the same timestamp $t_s$ with the target image, as the temporal reference since it can provide the sharp texture information.

\subsubsection{Event-based Low-light Image Enhancement}
A critical challenge in frame-based LIE methods arises from the inherent difficulty in disentangling edge-texture details from noise in short-exposure low-light images, as both components are high-frequency signatures \cite{zhu2020eemefn,xu2023low}. Taking advantage of event cameras in inherently embedding precise motions and sharp edges, recent E-LIE approaches explore reconstructing clear edges in dark areas \cite{liang2024towards}. However, they still struggle with the color distortion since E-LIE is an ill-posed problem. Based on this, we form the \textit{Enhancement} path, introducing the long-exposure image $I_l$ as the intensity reference,
\begin{equation}
    \hat{L}_s=\operatorname{E-LIE}(I_s, I_l, \mathcal{E}_{[t_s-\Delta t,t_s+\Delta t]}),
    \label{eq:elie}
\end{equation}
where $\hat{L}_s$ indicates the restoration results of $\operatorname{E-LIE}(\cdot)$, the event-based low-light image enhancement function, with the temporally surrounding events $\mathcal{E}_{[t_s-\Delta t,t_s+\Delta t]}$ as the edge and texture guidance. 

Ideally, we can adopt two dedicated modules for the above two sub-tasks, respectively, with the subsequent output fusion. However, degradations of reference information may lead to suboptimal performance: For E-MD in \cref{eq:emd}, edges and textures of the temporal reference $I_s$ are disturbed by the low intensity and annoying noise. And for E-LIE in \cref{eq:elie}, the motion blur and pixel displacement of the intensity reference $I_l$ lead to serious artifact results. 

\subsection{Dual-Path Parallel Architecture}
\label{sec:backbone}

\revision{To address the above-coupled problem, we design \myname as a dual-path parallel network to progressively restore high-quality images from dual-exposure image pairs and corresponding event streams. As illustrated in \cref{fig:pipeline}\subcref{(a)}, the key architecture of \myname is a combination of two parallel encoder-decoder structures, based on the UNet \cite{ronneberger2015u}. The two parallel feature propagation paths respectively implement the generic mappings defined in \cref{eq:emd,eq:elie}. Specifically, the \textit{Deblurring} path receives the long-exposure image $I_l$ and events $\mathcal{E}_{[t_s,t_e]}$ as inputs and outputs the deblurred image $\hat{L}_l$, while the \textit{Enhancement} path is fed with the short-exposure image $I_s$ and surrounding events $\mathcal{E}_{[t_s-\Delta t,t_s+\Delta t]}$ to produce the enhanced image $\hat{L}_s$. Moreover, we design the Dual-path Feature Alignment and Fusion (DFAF) module to implicitly realize the spatial alignment and effective interaction of features from the two paths. Our network design has two benefits: 
\begin{itemize}
    \item The parallel architecture decouples the complex orignal task and simplifies the learning of the two sub-tasks. It is an interpretable network that can improve the reconstruction performance of each sub-task.
    \item Through the DFAF module, features of the two paths can interact with each other at multiple stages, which can effectively mitigate the negative impact of the degraded reference information. 
\end{itemize}}

\revision{Finally, a Complementary Gated Fusion (CGF) module is designed to fuse the results and features from two paths for the final result $\hat{L}$ as,
\begin{equation}
    \hat{L}=\operatorname{CGF}(\hat{L}_l, \hat{L}_s, F'_l, F'_s),
\end{equation} 
where $F'_l, F'_s$ are the features of the \textit{Deblurring} and \textit{Enhancement} paths before the last convolution layers.}

\begin{figure}[t]
    \centering
    \includegraphics[width=0.8\linewidth]{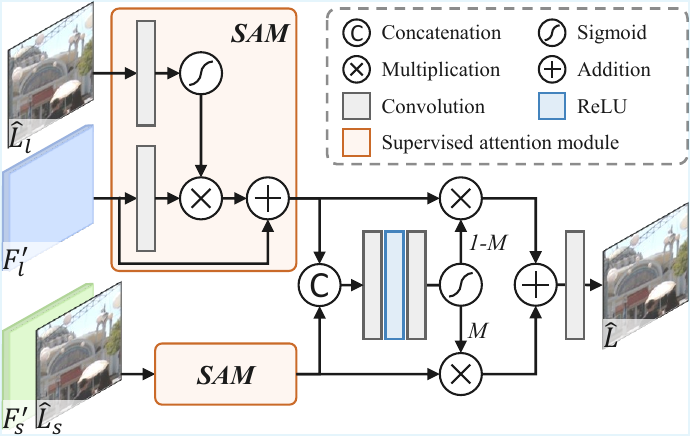}
    \caption{\revision{Details of the proposed Cross-Guided Fusion (CGF) module, which is designed to fuse the results and features of the \textit{Deblurring} and \textit{Enhancement} paths for the final result.}}
    \label{fig:cgf}
\end{figure}

\subsection{Dual-Path Feature Alignment and Fusion}
\label{sec:network}
\revision{Differing from current event-based single-exposure alignment \cite{xiao2024event,xiao2025event,tulyakov2021time} or multi-exposure fusion \cite{shaw2022hdr,messikommer2022multi,samra2023high,guo2024event} methods, our DFAF module purposes on handling the coupled degradation the coupled degradation caused by cross-exposure image differences and the spatial misalignment in dynamic scenes.}

We first define $F_l$/$F_s$ as the features of the \textit{Deblurring}/\textit{Enhancement} paths, which are aligned and fused gradually via the proposed Dual-path Feature Alignment and Fusion (DFAF) modules. We first extract the features $F_l,F_s$ respectively from the dual-exposure image pairs $I_s,I_l$ with the corresponding events $\mathcal{E}_{[t_s,t_e]},\mathcal{E}_{[t_s-\Delta t,t_s+\Delta t]}$ as,
\begin{equation}
    \begin{aligned}
        F_l&=\operatorname{Res}(\operatorname{Conv}(I_l,\mathcal{E}_{[t_s,t_e]})), \\
        F_s&=\operatorname{Res}(\operatorname{Conv}(I_s,\mathcal{E}_{[t_s-\Delta t,t_s+\Delta t]})),
    \end{aligned}
\end{equation}
where $\operatorname{Res}(\operatorname{Conv}(\cdot))$ means a sequential combination of a $3\times3$ convolution layer and a residual convolution layer. As depicted in \cref{fig:pipeline}\subcref{(b)}, our DFAF module contains two components, \ie, the deformable alignment and cross-attention fusion blocks. 

\subsubsection{Deformable Alignment Block}
The key target of our Deformable Alignment (DA) block is to implicitly solve the \cref{eq:edei_d1}, mapping the spatial domain of the long-exposure image feature to align that of the short-exposure image feature. Specifically, the features $F_l, F_s$ of the \textit{Deblurring} and \textit{Enhancement} paths are concatenated to calculate the learnable offset map $\Theta_{l \rightarrow s}$ via a convolution layer. Then $\Theta_{l\rightarrow s}$ is fed to the deformable convolution layer \cite{zhu2019deformable} for the spatial alignment between $F_l$ and $F_s$ as,
\begin{equation}
    \begin{aligned}
        \Theta_{l\rightarrow s}&=\operatorname{Conv}([F_l; F_s]), \\
        F'_l&=\operatorname{DCN}(F_l | \Theta_{l\rightarrow s}),\\
    \end{aligned}
    \label{eq:da}
\end{equation}
where \revision{$[\cdot;\cdot]$ means the concatenation} and $\operatorname{DCN}(\cdot)$ indicates the deformable convolution function and $F'_l$ is the aligned feature mapping from the spatial domain of $I_l$ to that of $I_s$.

\subsubsection{Cross-Attention Fusion Block}
\revision{Subsequently, we adopt the Cross-Attention Fusion (CAF) block to realize the \cref{eq:fdei_d1} to effectively integrate the complementary information of the aligned feature $F'_l$ and the feature of the \textit{Enhancement} path $F_s$. The CAF block first applys the channel-wise cross-attention fusion mechanism between $F'_l$ and $F_s$, and then enhances the features via the Feed-Forward Network (FFN) \cite{zamir2022restormer}. The overall process can be formulated as,
\begin{equation}
    \begin{aligned}
        F'_s&=\operatorname{Atten}(F_s, F'_l) + F_s, \\
        F'_s&=\operatorname{FFN}(F'_s) + F'_s,
    \end{aligned}
\end{equation}
where $F'_s$ is the enhanced feature and $\operatorname{Atten}(\cdot)$ is the channel-wise cross-attention fusion function with the multi-head mechanism. Specifically, the cross-attention fusion mechanism first calculates queries ($Q_s$) from the $F_s$ and keys ($K_l$) and values ($V_l$) from the $F'_l$ by the layer normalization and $1\times1$ depth-wise convolution layers. Then, we compute the channel-wise attention between vectorized features from the two paths. The above process can be formulated as,
\begin{equation}
    \begin{aligned}
        \operatorname{Atten}(F_s, F'_l)&=V_l\operatorname{Softmax}(\frac{Q^T_sK_l}{\sqrt{d}}), \operatorname{with} \\
        Q_s&=\operatorname{Conv}(\operatorname{LN}(F_s)), \\
        K_l, V_l&=\operatorname{Conv}(\operatorname{LN}(F'_l)),
    \end{aligned}
\end{equation}
where $\operatorname{LN}(\cdot)$ is the layer normalization function and $d$ is the dimension of the feature channel. FFN is implemented by two linear layers, one $3\times3$ convolution layers, and a GELU activation layer.}

Note that existing methods \cite{zhang2022unifying,yang2024motion} have proved that the information provided by long-exposure images and event streams is sufficient to solve the E-MD task. \revision{Hence, in the DA block, the feature of the \textit{Enhancement} path $F_s$ is only used as the temporal reference to guide the spatial alignment, not directly participating in the feature forward propagation of the \textit{Deblurring} path. In contrast, in the CAF block, the intensity information provided by the long-exposure image is necessary for the high color fidelity requirement of the reconstructed image. Thus, we need to fuse $F_l$ and $F_s$ in the feature forward propagation of the \textit{Enhancement} path. Such a module design allows the two paths to focus on their respective sub-tasks and improve the convergence ability of our \myname network.}

\subsection{\revision{Complementary Gated Fusion}}
\label{sec:cgf}
\revision{The Complementary Gated Fusion (CGF) module aims at fully utilizing the learned features and results of the two paths for the final reconstruction. As shown in \cref{fig:cgf}, the CGF module contains two components, \ie, the Supervised Attention Module (SAM) \cite{zamir2021multi} and the spatial attention-based fusion block. The SAM has been proved to be effective in guiding the network to focus on the important regions for image restoration. Thus, we adopt the SAM to enhance the features of the two paths before the final fusion. Specifically, we feed the features $F'_l$ and $F'_s$ and corresponding results $\hat{L}_l$ and $\hat{L}_s$ into two parallel SAM modules, respectively, as,
\begin{equation}
    \begin{aligned}
        F''_l = \operatorname{SAM}(\hat{L}_l, F'_l), \\
        F''_s = \operatorname{SAM}(\hat{L}_s, F'_s),
    \end{aligned}
\end{equation}
where $F''_l$ and $F''_s$ are enhanced features of the \textit{Deblurring} and \textit{Enhancement} paths via the SAM modules, respectively. Then, we calculate the spatial attention map $M$ between the two paths as,
\begin{equation}
    M = \operatorname{Sigmoid}(\operatorname{SA}([F''_l; F''_s])),
\end{equation}
where $\operatorname{SA}(\cdot)$ is the spatial attention function, which is implemented by two convolution layers and a ReLU operation. Finally, we fuse the results and features of the two paths with the calculated attention map as,
\begin{equation}
    \hat{L} = \operatorname{Conv}(M\cdot F''_s + (1-M)\cdot F''_l) + \hat{L}_s.
\end{equation}}

\begin{figure*}
    \begin{tabular}{cc}
        \hspace{-3mm}
        \includegraphics[width=0.3425\textwidth]{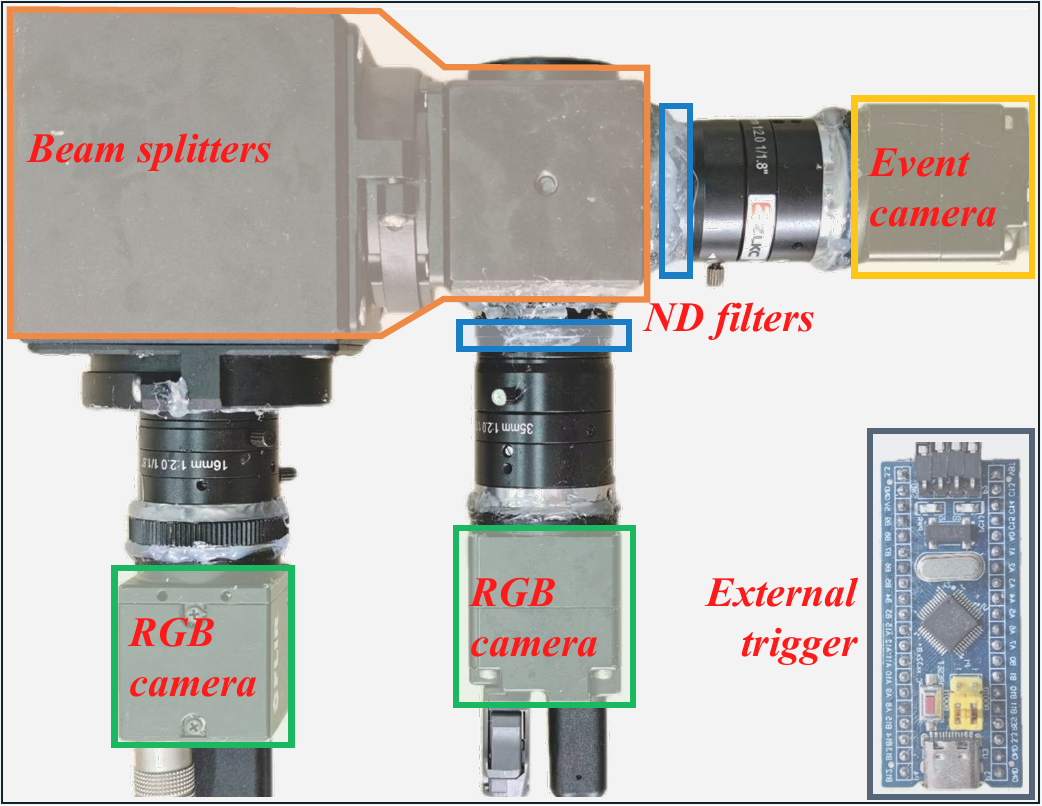} \hspace{-5mm} & \includegraphics[width=0.65\textwidth]{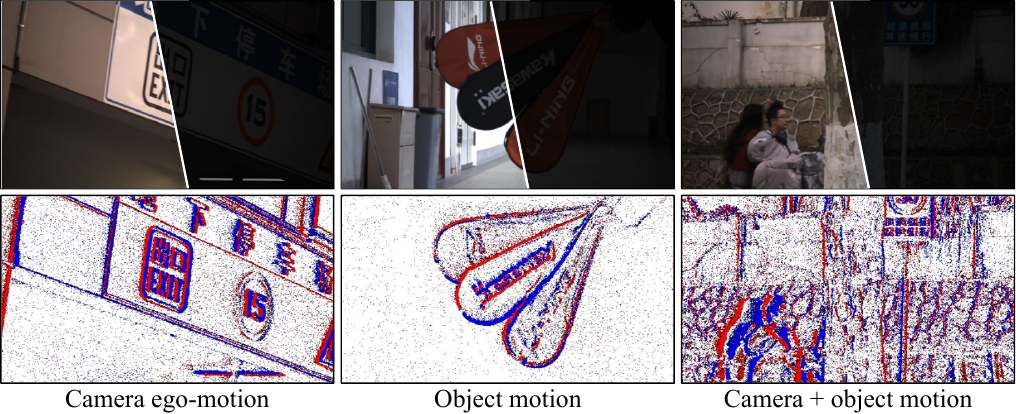} \\
        {\small (a) Hybrid camera system} & (b) {\small Scene examples in diverse motion patterns}
    \end{tabular}
    \caption{(a) The illustration of a beam splitter-based hybrid camera system for building the \mydata dataset. (b) \mydata contains the normal- and low-light image pairs (upper row) and the corresponding events (bottom row) with three diverse motion patterns.}
    \label{fig:dataset}
\end{figure*}

\subsection{Loss Function}
\label{sec:loss}
\revision{Since all the \textit{Deblurring}, \textit{Enhancement} paths, and the CGF module can output the reconstructed results, we supervise our \myname network by measuring the pixel-wise $\ell_1$ loss between the predicted images $\hat{L}_s, \hat{L}_l, \hat{L}$ and the corresponding ground-truth high-quality image $L$ at time $t_s$ as,
\begin{equation}
    \mathcal{L}=\lambda_1 \|\hat{L}-L\|_1 + \lambda_2 \|\hat{L}_s-L\|_1 + \lambda_3 \|\hat{L}_l-L\|_1,
    \label{eq:loss}
\end{equation}
where $\lambda_1, \lambda_2, \lambda_3$ are the balancing parameters.}

\section{Dataset}
\subsection{Limitations of Existing Datasets}
Recent works \cite{liang2024towards,zhang2024sim,liu2024seeing} have collect low-light scene datasets containing real-world events, yet they still have two limitations: \textbf{a) Lack of low- and normal-light image pairs.} Zhang \etal \cite{zhang2024sim} build the EVFI-LL dataset containing low-light event streams and noisy images, but no corresponding normal-light noise-free images. Liu \etal \cite{liu2024seeing} collect the RLED dataset for the event-based video reconstruction task, capturing normal-light images and using an ND8 filter for low-light event streams. \textbf{b) Limited motion patterns.} Liang \etal \cite{liang2024towards} introduce the SDE dataset, including paired low- and normal-light images, where they utilize a robotic arm to ensure spatiotemporal alignment. However, they only capture static scenes without considering moving objects. \revision{It motivates us to build a co-optical axis hybrid camera system and capture a new Dataset consisting of Paired normal- and low-light sharp Images and corresponding Events (\mydata) for evaluating event-based imaging methods in real-world scenes.}

\begin{table}
    \centering
    \caption{Comparisons of our dataset with the existing event-based low-light datasets. LL/NL means low-/normal-light images}
    \begin{tabular}{l c c c c c}
        \hline
        \multirow{2}{*}{Dataset} & \multicolumn{2}{c}{Image} & \multirow{2}{*}{Resolution} & \multirow{2}{*}{FPS} & \multirow{2}{*}{Motion pattern} \\
        \cline{2-3}
         & LL & NL & & & \\
        % Dataset & LI & NL & Resolution & FPS & Motion pattern \\
        \hline
        SDE \cite{liang2024towards} & \ding{51} & \ding{51} & 346$\times$260 & $<$30 & Camera motion \\
        EVFI-LL \cite{zhang2024sim} & \ding{51} & \ding{55} & 784$\times$611 & 64 & Moving object \\
        RLED \cite{liu2024seeing} & \ding{55} & \ding{51} & 1120$\times$660 & 10,25 & \textbf{Camera+Object} \\
        Ours & \ding{51} & \ding{51} & \textbf{1280$\times$720} & \textbf{72} & \textbf{Camera+Object} \\
        \hline
    \end{tabular}
    \label{tab:dataset}
\end{table}

\subsection{System Setup}
As shown in \cref{fig:dataset}\subcref{(a)}, we design a hybrid camera system where two same RGB cameras (FLIR BFS-U3-32S4C-C, $2048 \times 1536$) and one event camera (Prophesee EVK5, $1280 \times 720$) are co-optical axis aligned with two 50:50 beam splitters. We adjust the irradiance intensity of one of the RGB cameras and the event camera to be 1/16 of that of the other RGB camera using two ND8 filters to simulate low-light photography scenarios. Our hybrid camera system provides the temporally synchronized low- and normal-light image pairs with a higher spatial resolution ($1280 \times 720$) at a higher frame rate (72 FPS) than the abovementioned datasets, as shown in \cref{tab:dataset}. It can be used to simultaneously evaluate multiple frame-/event-based low-light imaging methods, \eg, long-exposure motion deblurring, short-exposure image enhancement, and dual-exposure imaging algorithms, under the same scene.

\subsection{Spatiotemporal Alignment}
For the temporal synchronization, we use an STM32 development board to transmit an external trigger signal to the three cameras. For the spatial alignment, the geometrical connection of cameras with two 50:50 cube beam splitters leads to a minimal baseline. Further, we perform pixel-wise alignment among multiple cameras based on the homography calculated by extrinsic calibration. Hence, our hybrid camera system can simultaneously produce spatiotemporal aligned normal-light high-quality images, low-light noisy images, and corresponding event streams, unlike SDE \cite{liang2024towards} that have to repeatedly capture the same static scene for a sequence. Based on this system, we collect new real-world \mydata dataset having three types of combination of object and camera motion, \ie, camera ego-motion, object motion, and joint camera and object motion as shown in \cref{fig:dataset}\subcref{(b)}.

\begin{figure}[t]
    \centering
    \begin{tabular}{c c}
        \hspace{-1mm}\includegraphics[width=0.48\linewidth]{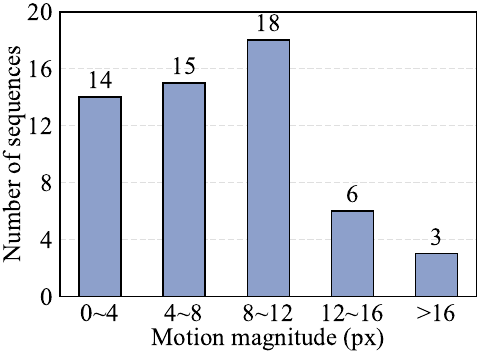}\hspace{-3mm} & \includegraphics[width=0.48\linewidth]{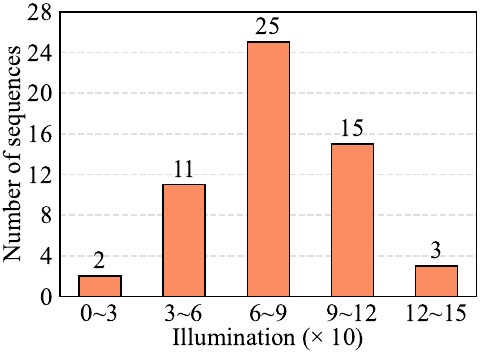} \\
        \revision{\small (a)} & \revision{\small (b)} \\
        \hspace{-1mm}\includegraphics[width=0.48\linewidth]{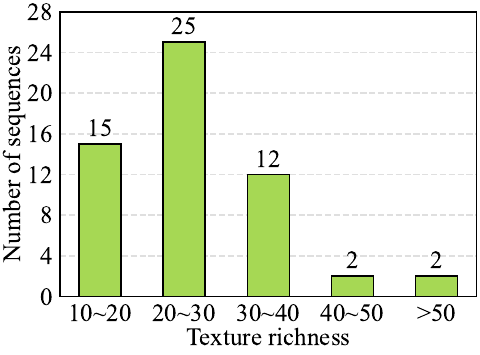}\hspace{-3mm} & \includegraphics[width=0.48\linewidth]{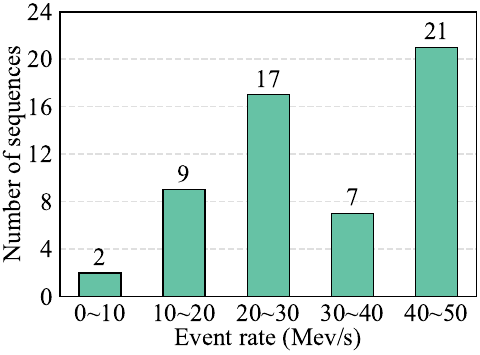} \\
        \revision{\small (c)} & \revision{\small (d)} \\
    \end{tabular}
    \caption{\revision{Statistical properties of the \myname dataset. Distributions of (a) Motion magnitude, (b) Illumination, (c) Texture richness, and (d) Event firing rate.}}
    \label{fig:dataset_statistic}
\end{figure}

\subsection{\revision{Statistical Property}}
\revision{\cref{fig:dataset_statistic} presents the statistical properties of the proposed \mydata dataset, including the distributions of: \textbf{1) Motion magnitude.} We adopt an OpenCV function, which uses the method proposed in \cite{farneback2003two}, to calculate the average pixel-wise optical flow magnitude between the consecutive normal-light images via the Farneb{\"a}ck method. The motion magnitude can reflect the severity of motion blur in the long-exposure image and the misalignment between the long- and short-exposure images. \textbf{2) Illumination variation.} The color space of normal-light images are firstly transformed from RGB to YCbCr. Then, we calculate the average pixel intensity of the Y channel as an indicator of lighting changes. \textbf{3) Texture richness.} We use the Sobel operator to calculate the average of the magnitudes of image edges and gradients to approximate the texture richness of scenes. \textbf{4) Event firing rate.} It is calculated by the number of millions of events per second (Mev/s).}

\section{Experiments}
In this section, we first introduce the datasets and implementation details in \cref{sec:setup}, and then evaluate the performance of our algorithm against state-of-the-art methods on synthetic and real-world datasets in \cref{sec:comparison}. Ablation study on the effectiveness of network architecture and the balancing parameter is presented in \cref{sec:ablation}. 

% Finally, we elucidate the complexity and runtime analysis in \cref{sec:runtime}.

\subsection{Experimental Setup}
\label{sec:setup}
\subsubsection{Datasets} Two different datasets containing dual-exposure image pairs, corresponding event streams, and the ground-truth images are employed in our experiments for network training and evaluation.

% \textbf{\textit{REDS-LL.}} We build a synthetic dataset upon REDS \cite{nah2019ntire}. We follow \cite{lv2021attention,zhang2021learning} for the synthesis of short-exposure images $I_s$. We first lower the pixel intensity to generate the low-light images without noise from the normal-light ones $L$ using gamma correction and linear scaling, and then add noise with the signal-dependent Gaussian distribution as,
% \begin{equation}
%     \begin{aligned}
%         I_s&\sim\mathcal{N}(D, \sigma_pD+\sigma^2_g), \operatorname{with} \\
%         D&=\beta\times(\alpha\times L)^\gamma,
%     \end{aligned}
%     \label{eq:Is_L}
% \end{equation}
% where $\alpha,\beta,\gamma$ are sampled from the uniform distribution $\mathcal{U}(0.9,1),\mathcal{U}(0.5,1),\mathcal{U}(2,3.5)$, respectively. $\mathcal{N}$ is the Gaussian distribution, and $\sigma_p$ and $\sigma_g$ are the noise parameters that are sampled from the uniform distribution $\mathcal{U}(0.05,0.1)$. For each sequence, we first generate high frame-rate videos by interpolating 7 images between consecutive frames with RIFE \cite{huang2022real} and then synthesize long-exposure images $I_l$ by averaging 49 sharp images of the high frame-rate videos. Events are generated via v2e \cite{hu2021v2e} with the cut-off frequency set at 15 Hz and the temporal noise rate at 1 Hz to approximate the event degradation in real-world low-light scenarios.

\textbf{\textit{REDS-LL.}} We build a synthetic dataset upon REDS \cite{nah2019ntire}. We lower the pixel values for frames for the synthesis of short-exposure images $I_s$ and add noise as \cref{eq:i_s}, $\alpha,\beta,\gamma$ are sampled from the uniform distribution $\mathcal{U}(0.9,1),\mathcal{U}(0.5,1),\mathcal{U}(2,3.5)$, respectively, and $\sigma_p$ and $\sigma_g$ are sampled from the uniform distribution $\mathcal{U}(0.05,0.1)$. For each sequence, we first generate high frame-rate videos by interpolating 7 images between consecutive frames with RIFE \cite{huang2022real} and then synthesize long-exposure images $I_l$ by averaging 49 sharp images of the high frame-rate videos. Events are generated via v2e \cite{hu2021v2e} with the cut-off frequency set at 15 Hz and the temporal noise rate at 1 Hz to approximate the event degradation in real-world low-light scenarios.

\textbf{\textit{\mydata.}} The captured real-world normal- and low-light images work as the ground-truth high-quality images and low-light images, respectively. Further, the long-exposure images are synthesized from the normal-light images following the same pipeline as REDS-LL. Meanwhile, our \mydata records corresponding event streams at the same spatial resolution as images. We divide \mydata into two parts, \ie, 40 sequences as the training set and 16 sequences as the testing set. The training and testing sets have 2044 and 573 short- and long-exposure image pairs, respectively.

\revision{Similar to existing DEI methods \cite{mustaniemi2020lsd2,chang2021low,zhao2022d2hnet}, our \myname network is based on the assumption that long-exposure images suffer only from motion blur without brightness degradation, while short-exposure images have sharp textures but are affected by noise and color distortion. Thus, the above datasets are synthesized and captured under the scenarios where the above assumption holds. In the \cref{sec:limitation}, we discuss the failure cases when the above assumption is broken.}

\begin{table*}
    \centering
    \caption{Quantitative comparisons of the proposed \myname to the state-of-the-art methods on the REDS-LL and \mydata datasets. \revision{The best performance is highlighted in \textbf{bold} and second best is \underline{underlined}.}}
    \begin{tabular}{l l c c c c c c}
        \hline
        \multirow{2}{*}{Method} && \multirow{2}{*}{Event} & \multicolumn{2}{c}{REDS-LL} && \multicolumn{2}{c}{\mydata} \\
        \cline{4-5} \cline{7-8}
         & & & PSNR $\uparrow$ & SSIM $\uparrow$ && PSNR $\uparrow$ & SSIM $\uparrow$ \\
        \hline
        \multirow{4}{*}{Low-light image enhancement} & Uretinex-Net \cite{wu2022uretinex} & \ding{55} & 25.621 & 0.9020 && 25.829 & 0.8613 \\
         & EMNet \cite{ye2023glow} & \ding{55} & 30.317 & 0.9253 && 26.776 & 0.8982 \\
         & EvLowLight \cite{liang2023coherent} & \ding{51} & 25.519 & 0.9217 && 26.808 & 0.8666 \\
         & EvLight \cite{liang2024towards} & \ding{51} & 31.211 & 0.9375 && 28.203 & 0.9102 \\
        \hline
        \multirow{3}{*}{Motion deblurring} & MSSNet \cite{kim2022mssnet} & \ding{55} & 30.181 & 0.8883 && 28.890 & 0.8538 \\
         & EFNet \cite{sun2022event} & \ding{51} & 32.056 & 0.9085 && 30.799 & 0.8907 \\
         & STCNet \cite{yang2024motion} & \ding{51} & 31.693 & 0.9020 && 30.030 & 0.8818 \\
         \hline
        \multirow{5}{*}{Dual-exposure imaging} & LSD$_2$ \cite{mustaniemi2020lsd2} & \ding{55} & 34.079 & 0.9470 && 28.965 & 0.8899 \\
         & LSFNet \cite{chang2021low} & \ding{55} & 34.444 & 0.9492 && 29.100 & 0.8992 \\
         & D2HNet \cite{zhao2022d2hnet} & \ding{55} & 35.529 & \underline{0.9560} && 30.851 & 0.9302 \\
         & \myname ($\hat{L}_s$) & \ding{51} & \underline{37.895} & \textbf{0.9695} && \underline{32.255} & \underline{0.9469} \\
         & \revision{\myname ($\hat{L}$)} & \ding{51} & \revision{\textbf{37.904}} & \revision{\textbf{0.9695}} && \revision{\textbf{32.263}} & \revision{\textbf{0.9473}} \\
        \hline
    \end{tabular}
    \label{tab:quanti}
\end{table*}

% In \cref{sec:lambda}, we will discuss the impact of different values of $\lambda$.

\begin{figure*}
    \centering
    \includegraphics[width=0.975\textwidth]{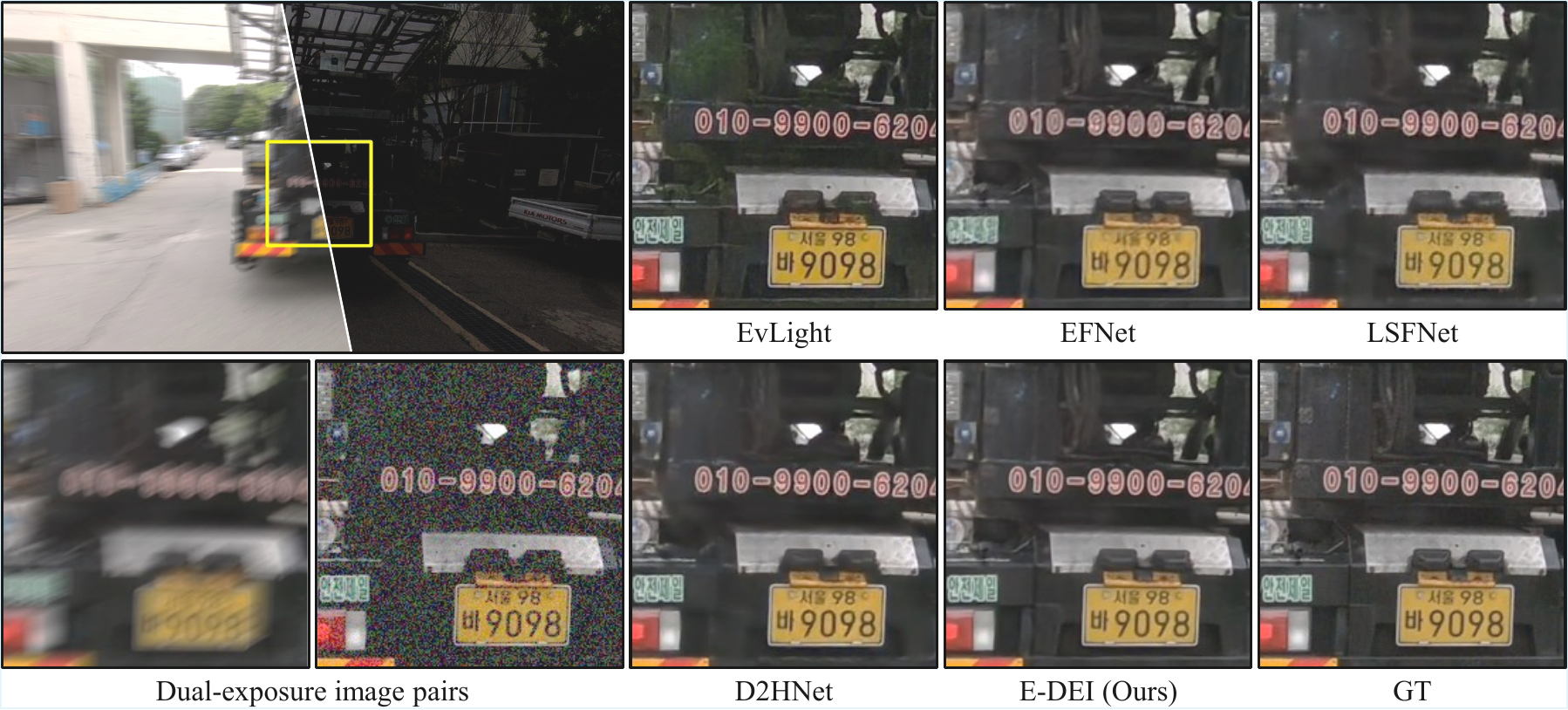}\\
    \includegraphics[width=0.975\textwidth]{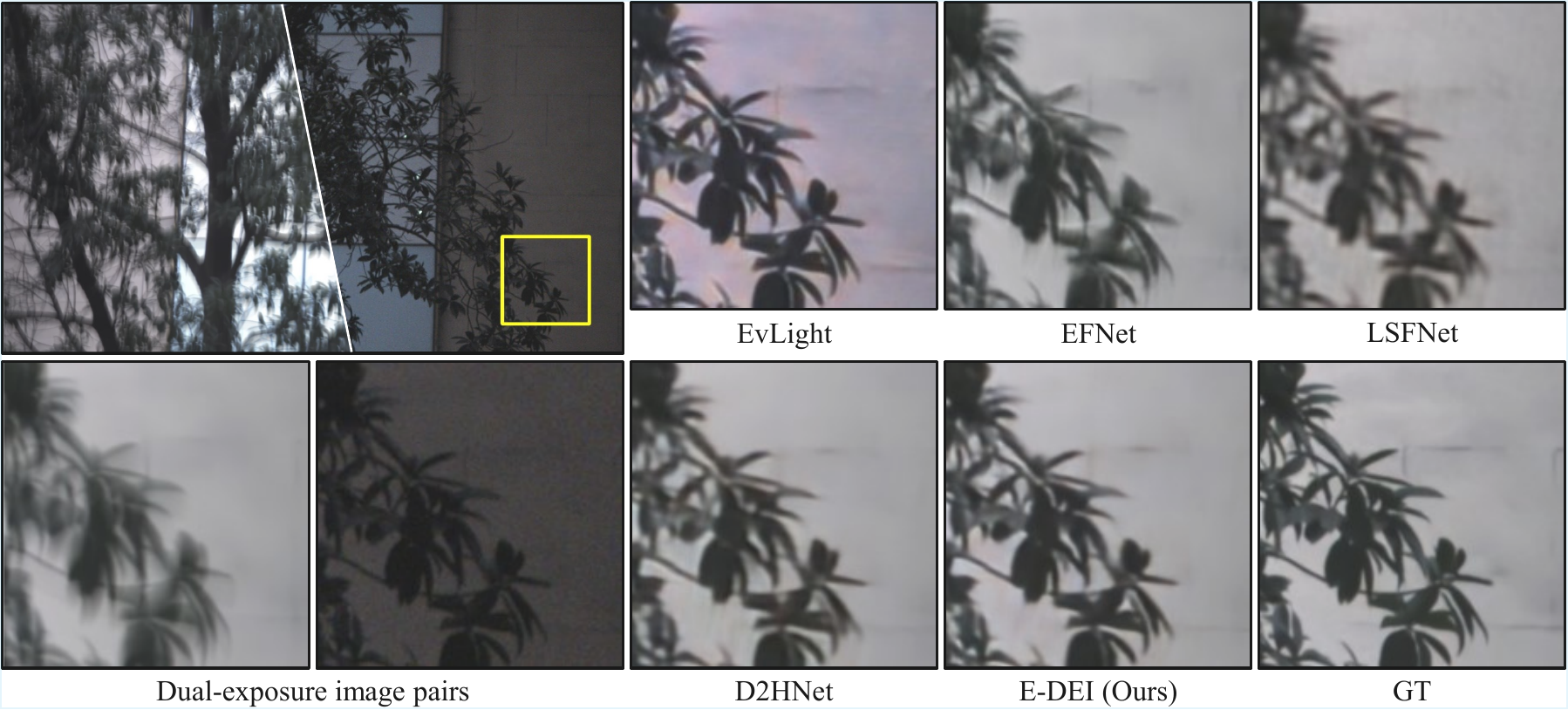}
    \caption{Qualitative comparisons with state-of-the-art methods on the REDS-LL (upper low) and \mydata (lower row) datasets. For a better view, details are zoomed in and the brightness of short-exposure images is rescaled.}
    \label{fig:exp_reds_ried}
\end{figure*}

\subsubsection{Implementation Details}
Our network is implemented using Pytorch and trained on a single NVIDIA GeForce RTX 3090 GPU with a batch size of 8 by default. During training, we randomly crop the samples into $256 \times 256$ patches. Adam optimizer is adopted, accompanied with the SGDR schedule. \revision{In the implementation, we train our \myname network in two stages. Firstly, we train the two parallel interactivate paths with balancing parameters $[\lambda_1,\lambda_2,\lambda_3]=[0,1,0.5]$. Secondly, we fix the parameters of the two paths and only train the CGF module with $[\lambda_1,\lambda_2,\lambda_3]=[1,0,0]$. The maximum epoch of training iterations is set to 100 in the first stage and 50 in the second stage for the REDS-LL dataset, while it is set to 200 and 100 for the \mydata dataset, respectively. The learning rate starts at $1\times10^{-4}$ in the first stage and $5\times10^{-5}$ in the second stage, respectively.} 

% \begin{figure*}
%     \centering
%     \includegraphics[width=0.975\textwidth]{figs/result/reds_005_483.png}\\
%     \includegraphics[width=0.975\textwidth]{figs/result/reds_015_081.png}
%     \caption{Qualitative comparisons with state-of-the-art methods on the REDS-LL dataset. For a better view, details are zoomed in and the brightness of short-exposure images is rescaled.}
%     \label{fig:exp_reds}
% \end{figure*}

% \begin{figure*}
%     \centering
%     \includegraphics[width=0.975\textwidth]{figs/result/ried_20_16_27_43_007.png} \\
%     \includegraphics[width=0.975\textwidth]{figs/result/ried_06_16_56_51_076.png}
%     \caption{Qualitative comparisons with state-of-the-art methods on the \mydata dataset. Details are zoomed in for a better view.}
%     \label{fig:exp_ried}
% \end{figure*}

\subsection{Qualitative and Quantitative Comparisons}
\label{sec:comparison}
We compare our algorithm with various methods based on three different image restoration strategies, \ie, low-light image enhancement methods applied on the short-exposure image (Uretinex-Net \cite{wu2022uretinex}, EMNet \cite{ye2023glow}, EvLowLight \cite{liang2023coherent}, EvLight \cite{liang2024towards}), motion deblurring methods for the long-exposure image (MSSNet \cite{kim2022mssnet}, EFNet \cite{sun2022event}, STCNet \cite{yang2024motion}), and the dual-exposure imaging methods (LSD$_2$ \cite{mustaniemi2020lsd2}, LSFNet \cite{chang2021low}, D2HNet \cite{zhao2022d2hnet}), in terms of Peak Signal to Noise Ratio (PSNR, higher is better) and Structural SIMilarity (SSIM, higher is better). We retrain all the above methods on the REDS-LL and \mydata datasets for a fair comparison. \revision{}

Quantitative results on the REDS-LL and \mydata datasets are reported in \cref{tab:quanti}. Motion deblurring approaches for long-exposure images generally attain higher PSNR than image enhancement methods for short-exposure images, due to their superior color fidelity retention. However, the short-exposure image enhancement methods often yield higher SSIM, as the short exposure setting helps to record sharper texture details in the dynamic scenes. Integrating the benefits of both long- and short-exposure imaging techniques can markedly elevate the overall image quality. \revision{Thanks to the introduced visual modality of event streams for effective and temporally consistent inter- and intra-frame motion estimation, the proposed dual-path parallel architecture of \myname outperforms the others in terms of all metrics by a large margin, demonstrating the effectiveness of our algorithm. The Complementary Gated Fusion (CGF) module can further improve the performance of our \myname by fully utilizing the learned features and results of the two paths, as evidenced by the performance improvement from $\hat{L}_s$ to $\hat{L}$ in \cref{tab:quanti}.}

\begin{table}
    \centering
    \caption{\revision{Ablation study on the feeding strategies of the event streams.}}
    \revision{\begin{tabular}{c c c c c}
        \hline
        \multirow{2}{*}{Ex.} & \multicolumn{2}{c}{Event} & \multirow{2}{*}{PSNR $\uparrow$} & \multirow{2}{*}{SSIM $\uparrow$} \\
        \cline{2-3}
         & \textit{Deblurring} & \textit{Enhancement} & & \\
        \hline
        1 & \ding{55} & \ding{55} & 34.522 & 0.9471 \\
        2 & \ding{51} & \ding{55} & 37.756 & 0.9684 \\
        3 & \ding{55} & \ding{51} & 36.281 & 0.9601 \\
        4 & \ding{51} & \ding{51} & \textbf{37.895} & \textbf{0.9695} \\
        \hline
    \end{tabular}}
    \label{tab:abla_event}
\end{table}

\begin{table}
    \centering
    \caption{\revision{Quantitative comparisons of the proposed \myname, which is a parallel architecture, to the serial pipeline design.}}
    \revision{\begin{tabular}{l c c c c c}
        \hline
        \multirow{2}{*}{Architecture} & \multicolumn{2}{c}{$\hat{L}_s$} && \multicolumn{2}{c}{$\hat{L}$}\\
        \cline{2-3} \cline{5-6}
        & PSNR$\uparrow$ & SSIM$\uparrow$ && PSNR$\uparrow$ & SSIM$\uparrow$ \\
        \hline
        Serial pipeline & 35.984 & 0.9581 && 36.702 & 0.9658 \\
        Parallel design (Ours) & \textbf{37.895} & \textbf{0.9695} && \textbf{37.904} & \textbf{0.9695} \\
        \hline
    \end{tabular}}
    \label{tab:abla_design}
\end{table}

Corresponding visual comparison results on the REDS-LL and \mydata datasets are shown in \cref{fig:exp_reds_ried}. Directly utilizing low-light noisy images to reconstruct normal-light results is ill-posed. It is evidenced in the results of EvLight \cite{liang2024towards}, which suffers from the serious color distortion. In motion deblurring methods for long-exposure images, compared to short-exposure inputs, the ample exposure time greatly boosts color fidelity. Further, event-based methods, \eg, EFNet \cite{sun2022event}, leverage intra-frame motion information from events and recover clearer results from severely texture-corrupted blurry images. However, the imaging results are still unsatisfactory due to inevitable artifacts, \eg, the blurry license plate in the upper row of \cref{fig:exp_reds_ried}. Conversely, frame-based dual-exposure imaging methods, \eg, LSFNet \cite{chang2021low} and D2HNet \cite{zhao2022d2hnet}, utilize the complementary information of long and short exposures, which can remove noise and blur simultaneously and restore true color. However, discrepancies in brightness, noise level, and texture sharpness caused by varying exposure times prevent frame-based methods from precisely compensating for spatial differences between image pairs, causing some smoothed details and artifacts, \eg, leaves in the lower row of \cref{fig:exp_reds_ried}. With the assistance of the continuous event streams, our \myname network can restore vivid textures and sharp edges.

\subsection{Ablation Study}
\label{sec:ablation}
\revision{In this subsection, we perform a diverse set of ablation studies to investigate event contributions, the importance of each module of the proposed network and the influence of the balancing parameter of the loss function.}

\begin{table}
    \centering
    \caption{\revision{Ablation study of the dual-path backbone and modules of our network on the REDS-LL dataset. Enc. and Dec. indicate the architecture of the encoder and decoder. DP and SP mean the dual- and single-path backbone.}}
    \revision{\begin{tabular}{c c c c c c c c}
        \hline
        \multirow{2}{*}{Ex.} & \multicolumn{2}{c}{Backbone} && \multicolumn{2}{c}{DFAF module} & \multirow{2}{*}{PSNR $\uparrow$} & \multirow{2}{*}{SSIM $\uparrow$} \\
        \cline{2-3} \cline{5-6}
         & Enc. & Dec. && DA & CAF & & \\
        \hline
        1 & SP & SP && \ding{55} & \ding{55} & 37.376 & 0.9657 \\
        2 & DP & SP && \ding{51} & \ding{51} & 37.637 & 0.9673 \\
        3 & DP & DP && \ding{55} & \ding{55} & 36.199 & 0.9591 \\
        4 & DP & DP && \ding{51} & \ding{55} & 35.876 & 0.9645 \\
        5 & DP & DP && \ding{55} & \ding{51} & 37.624 & 0.9674 \\
        6 & DP & DP && \ding{51} & \ding{51} & \textbf{37.895} & \textbf{0.9695} \\
        \hline
    \end{tabular}}
    \label{tab:abla_network}
\end{table}

\subsubsection{\revision{Contributions of Event Modality}} 
\revision{The high temporal resolution and and exposure-independent nature of the event data offer two advantages for the \myname task: 1) Supplementing lost inter- and intra-frame scene motion informationfor precise spatial alignment. 2) Providing texture cues for the short-exposure image, which can help to suppress noise and enhance texture details. To validate the contribution of event data, we conduct ablation studies of the feeding strategies on the $\hat{L}_s$. Quantitative results are shown in \cref{tab:abla_event}. Without the motion information extracted from events, the imaging results drastically decrease (the grass within the green box of \cref{fig:abla_network}). Feeding events into only one path can significantly improve the performance, especially for the \textit{Deblurring} path (Ex. 2), which demonstrates that the motion information from events is more crucial for the deblurring and alignment process. Moreover, feeding event data into both paths (Ex. 4) achieves the best performance, which demonstrates the complementary benefits of event data for both contributions.}

\def\ssxxsone{(-0.8,-0.8)} % 小方框位置
\def\ssyysone{(2.2,-2.1)} % 大方框位置
\def\ssxxstwo{(-1.45,0.89)} % 小方框位置
\def\ssyystwo{(-0.05,-2.1)} % 大方框位置
\def\ssmag{4} % 放大倍数
\def\sswspy{2.15cm} % 放大框宽度 4
\def\sshspy{1.6125cm} % 放大框高度 3
\def\sswidth{0.245\textwidth} %子图大小
\begin{figure*}[]
    \centering
    \begin{tabular}{c c c c}
        % \vspace{-0.5mm}
        \hspace{-3mm}
        \begin{tikzpicture}[spy using outlines={rectangle,green,magnification=\ssmag,width=\sswspy,height=\sshspy},inner sep=0]
    		\node {\includegraphics[width=\sswidth]{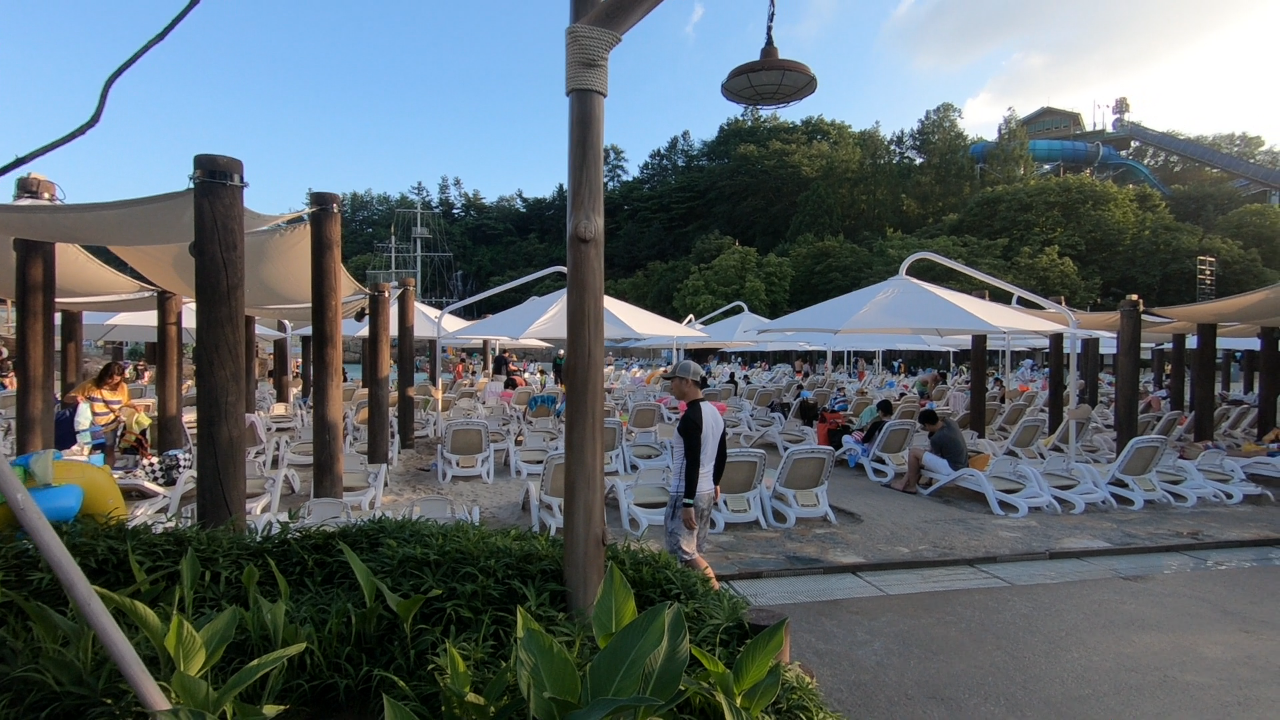}};
    		\spy on \ssxxsone in node [left] at \ssyysone;
            \spy [red] on \ssxxstwo in node [left,red] at \ssyystwo;
    	\end{tikzpicture} \hspace{-4.5mm} & 
        \begin{tikzpicture}[spy using outlines={rectangle,green,magnification=\ssmag,width=\sswspy,height=\sshspy},inner sep=0]
    		\node {\includegraphics[width=\sswidth]{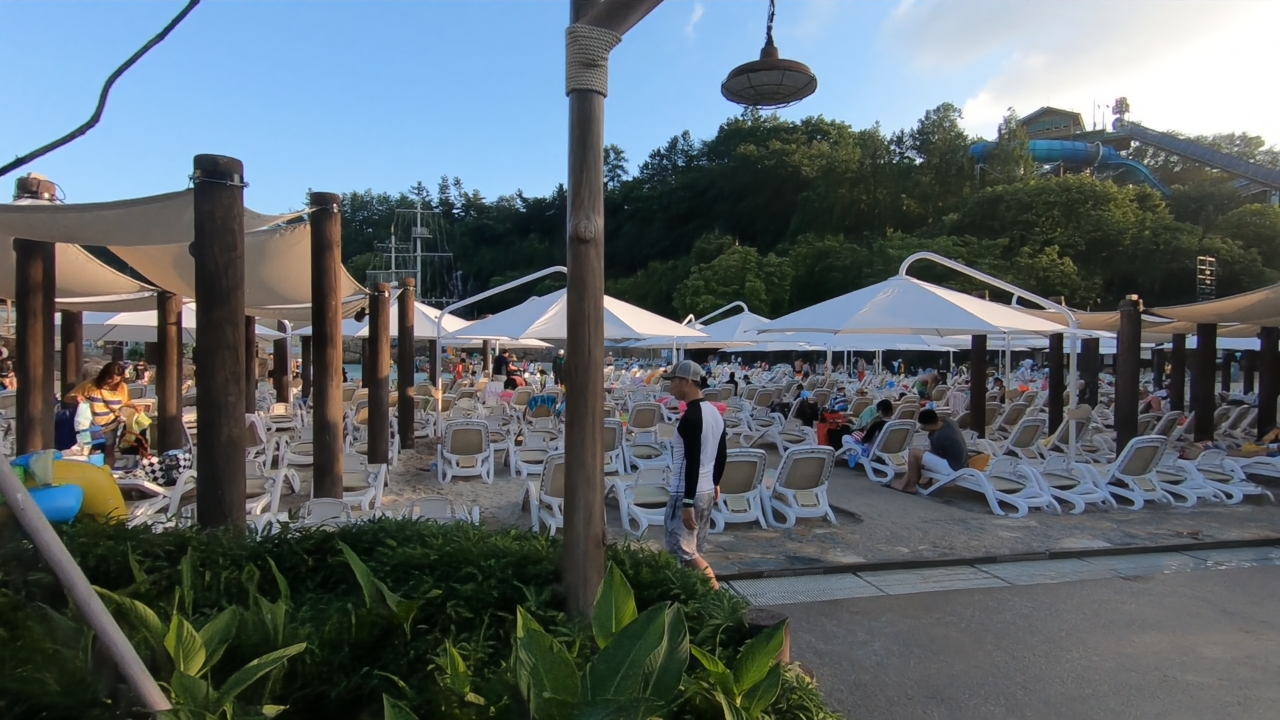}};
    		\spy on \ssxxsone in node [left] at \ssyysone;
            \spy [red] on \ssxxstwo in node [left,red] at \ssyystwo;
    	\end{tikzpicture} \hspace{-4.5mm} & 
        \begin{tikzpicture}[spy using outlines={rectangle,green,magnification=\ssmag,width=\sswspy,height=\sshspy},inner sep=0]
    		\node {\includegraphics[width=\sswidth]{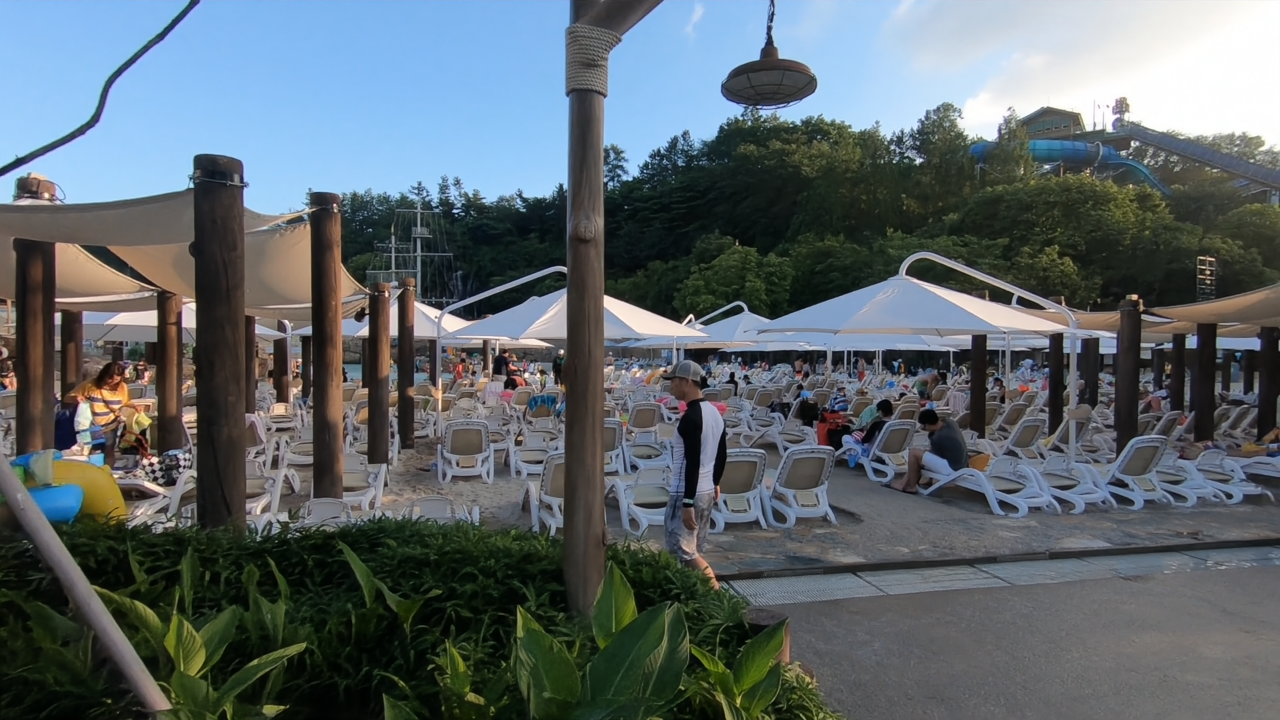}};
    		\spy on \ssxxsone in node [left] at \ssyysone;
            \spy [red] on \ssxxstwo in node [left,red] at \ssyystwo;
    	\end{tikzpicture} \hspace{-4.5mm} & 
        \begin{tikzpicture}[spy using outlines={rectangle,green,magnification=\ssmag,width=\sswspy,height=\sshspy},inner sep=0]
    		\node {\includegraphics[width=\sswidth]{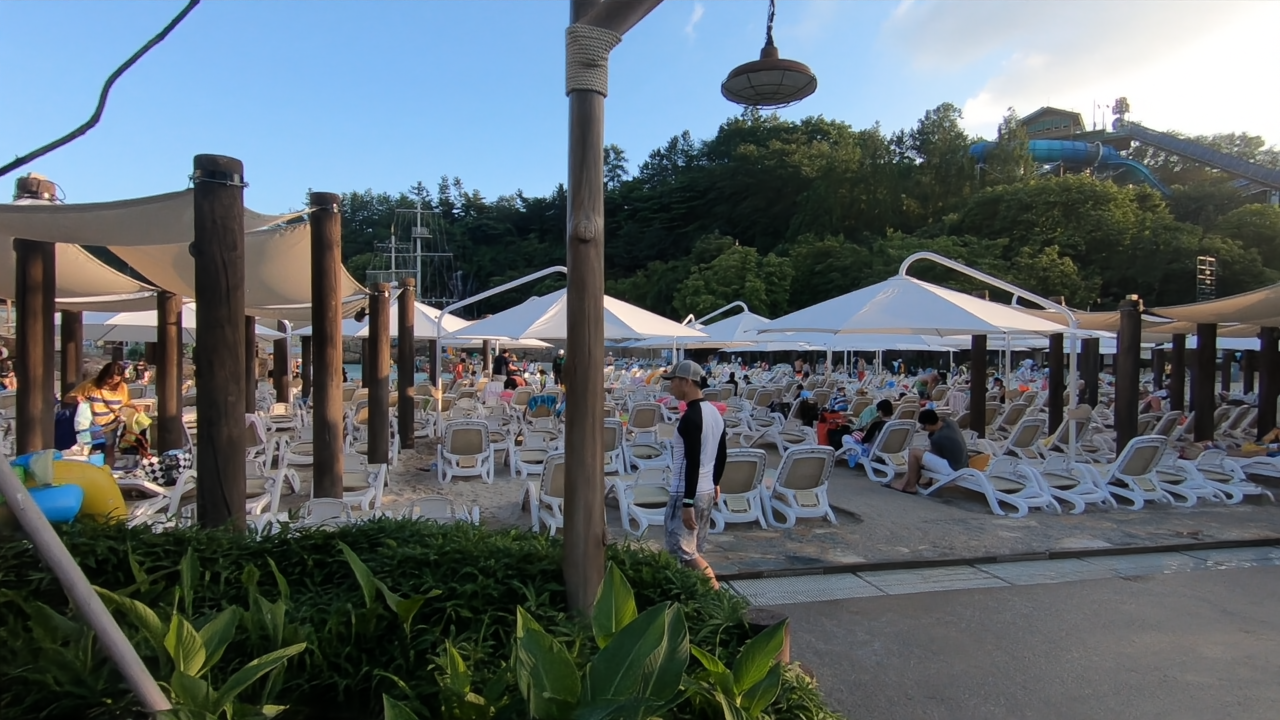}};
    		\spy on \ssxxsone in node [left] at \ssyysone;
            \spy [red] on \ssxxstwo in node [left,red] at \ssyystwo;
    	\end{tikzpicture} \vspace{-1mm} \\
        \hspace{-3mm}
        \begin{tikzpicture}[spy using outlines={rectangle,green,magnification=\ssmag,width=\sswspy,height=\sshspy},inner sep=0]
    		\node {\includegraphics[width=\sswidth]{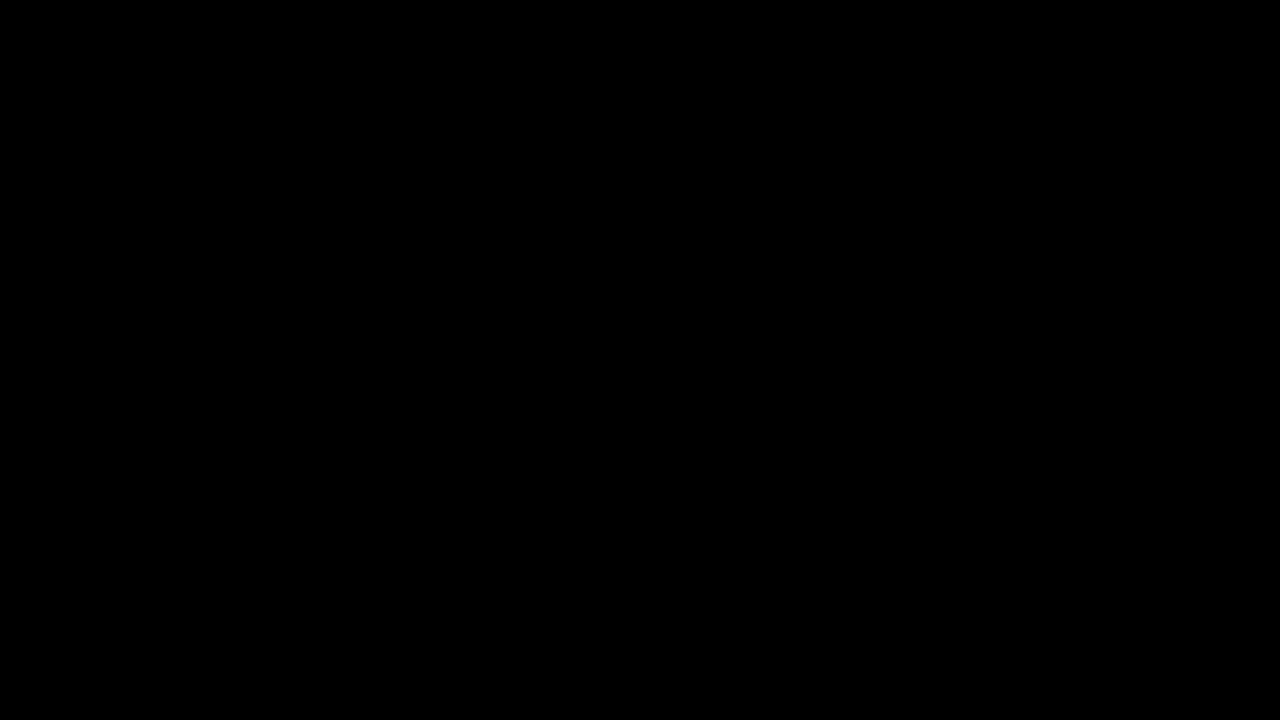}};
    		\spy on \ssxxsone in node [left] at \ssyysone;
            \spy [red] on \ssxxstwo in node [left,red] at \ssyystwo;
    	\end{tikzpicture} \hspace{-4.5mm} & 
        \begin{tikzpicture}[spy using outlines={rectangle,green,magnification=\ssmag,width=\sswspy,height=\sshspy},inner sep=0]
    		\node {\includegraphics[width=\sswidth]{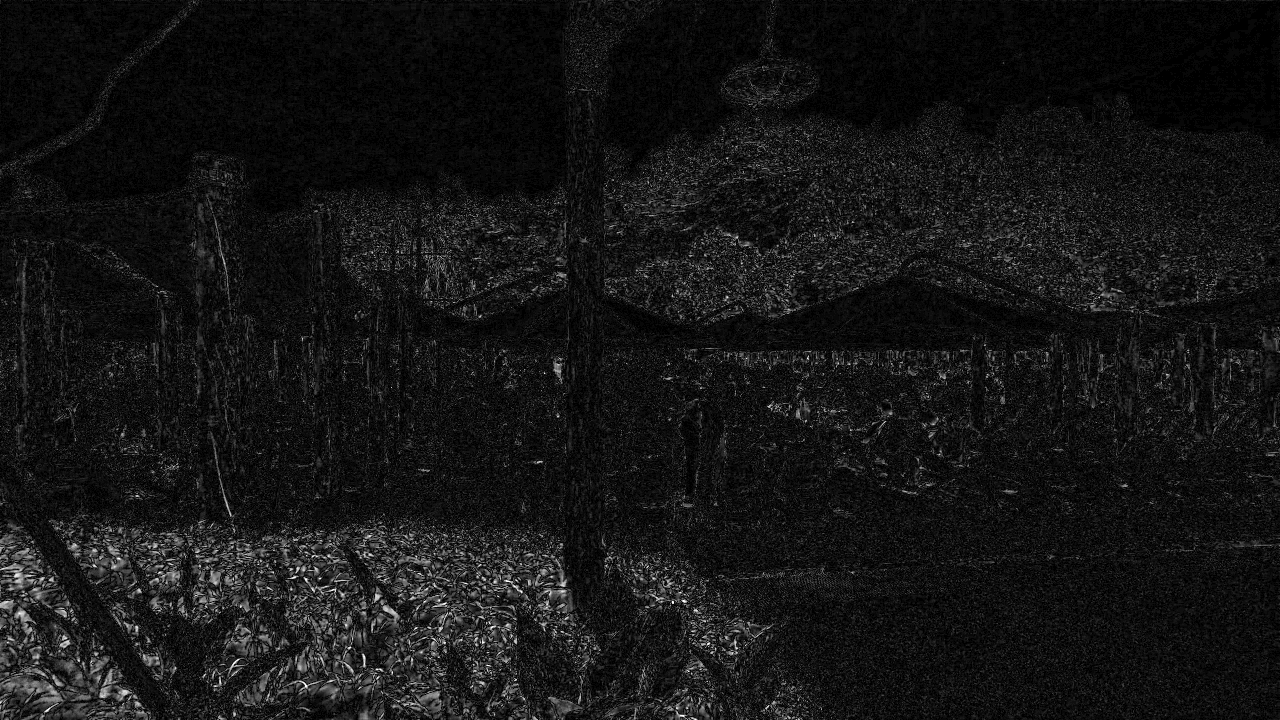}};
    		\spy on \ssxxsone in node [left] at \ssyysone;
            \spy [red] on \ssxxstwo in node [left,red] at \ssyystwo;
    	\end{tikzpicture} \hspace{-4.5mm} & 
        \begin{tikzpicture}[spy using outlines={rectangle,green,magnification=\ssmag,width=\sswspy,height=\sshspy},inner sep=0]
    		\node {\includegraphics[width=\sswidth]{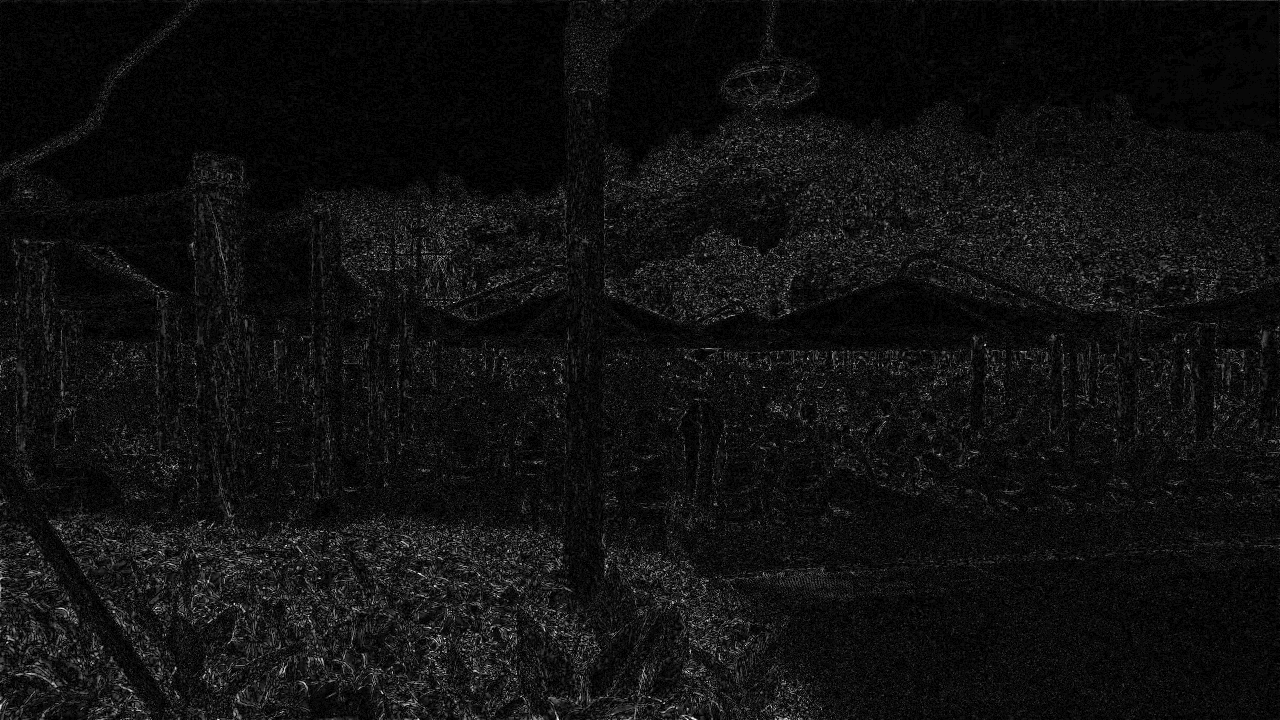}};
    		\spy on \ssxxsone in node [left] at \ssyysone;
            \spy [red] on \ssxxstwo in node [left,red] at \ssyystwo;
    	\end{tikzpicture} \hspace{-4.5mm} & 
        \begin{tikzpicture}[spy using outlines={rectangle,green,magnification=\ssmag,width=\sswspy,height=\sshspy},inner sep=0]
    		\node {\includegraphics[width=\sswidth]{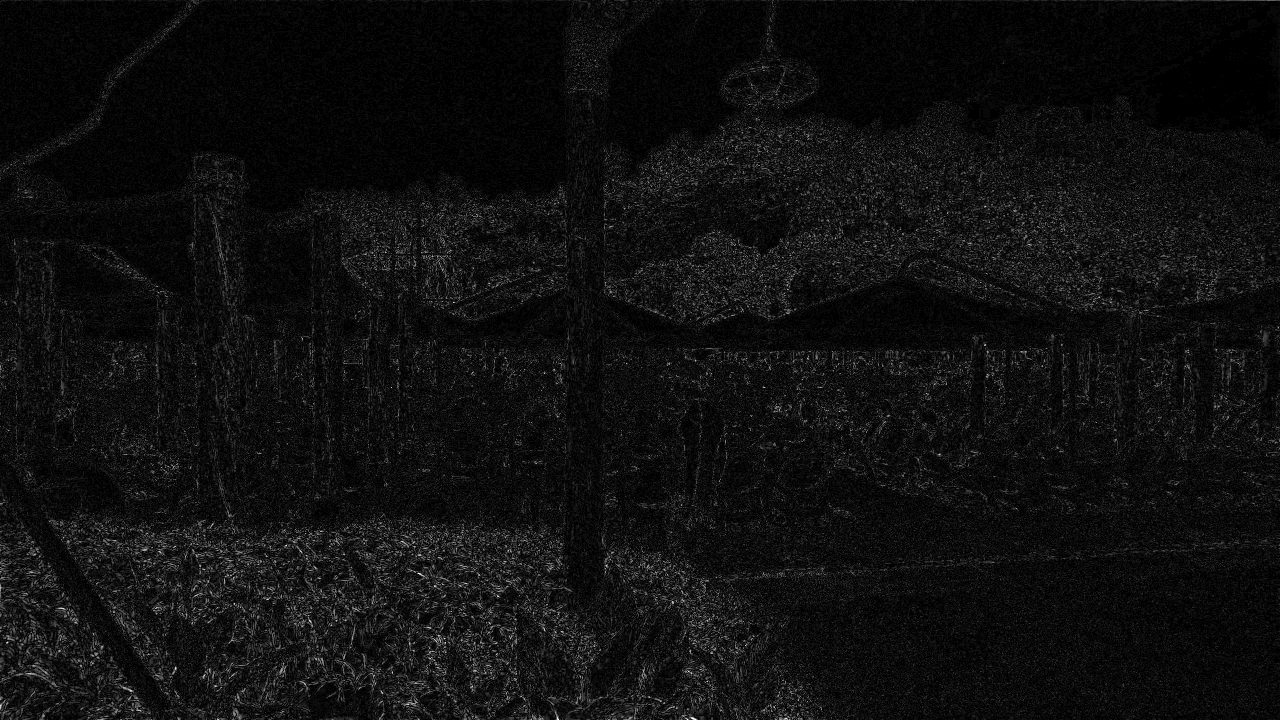}};
    		\spy on \ssxxsone in node [left] at \ssyysone;
            \spy [red] on \ssxxstwo in node [left,red] at \ssyystwo;
    	\end{tikzpicture} \vspace{-1mm} \\
        {\small GT} & {\small w/o Events input} & {\small w/ Single-path Enc. \& Dec.} & {\small w/ Single-path Dec.} \vspace{-1mm} \\
        {\small (PSNR$\uparrow$, SSIM$\uparrow$)} & {\small (35.214, 0.9413)} & {\small (37.178, 0.9649)} & {\small (37.678, 0.9680)} \\
        % \vspace{-0.5mm}
        \hspace{-3mm}
        \begin{tikzpicture}[spy using outlines={rectangle,green,magnification=\ssmag,width=\sswspy,height=\sshspy},inner sep=0]
    		\node {\includegraphics[width=\sswidth]{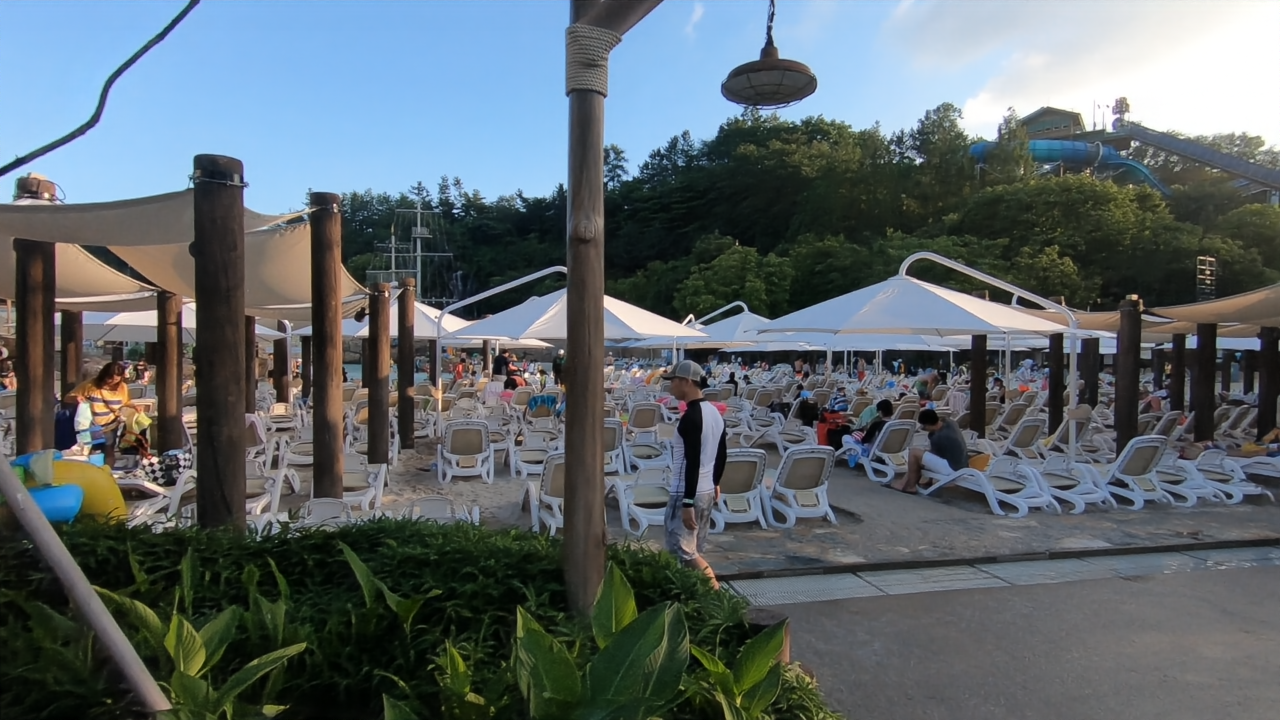}};
    		\spy on \ssxxsone in node [left] at \ssyysone;
            \spy [red] on \ssxxstwo in node [left,red] at \ssyystwo;
    	\end{tikzpicture} \hspace{-4.5mm} & 
        \begin{tikzpicture}[spy using outlines={rectangle,green,magnification=\ssmag,width=\sswspy,height=\sshspy},inner sep=0]
    		\node {\includegraphics[width=\sswidth]{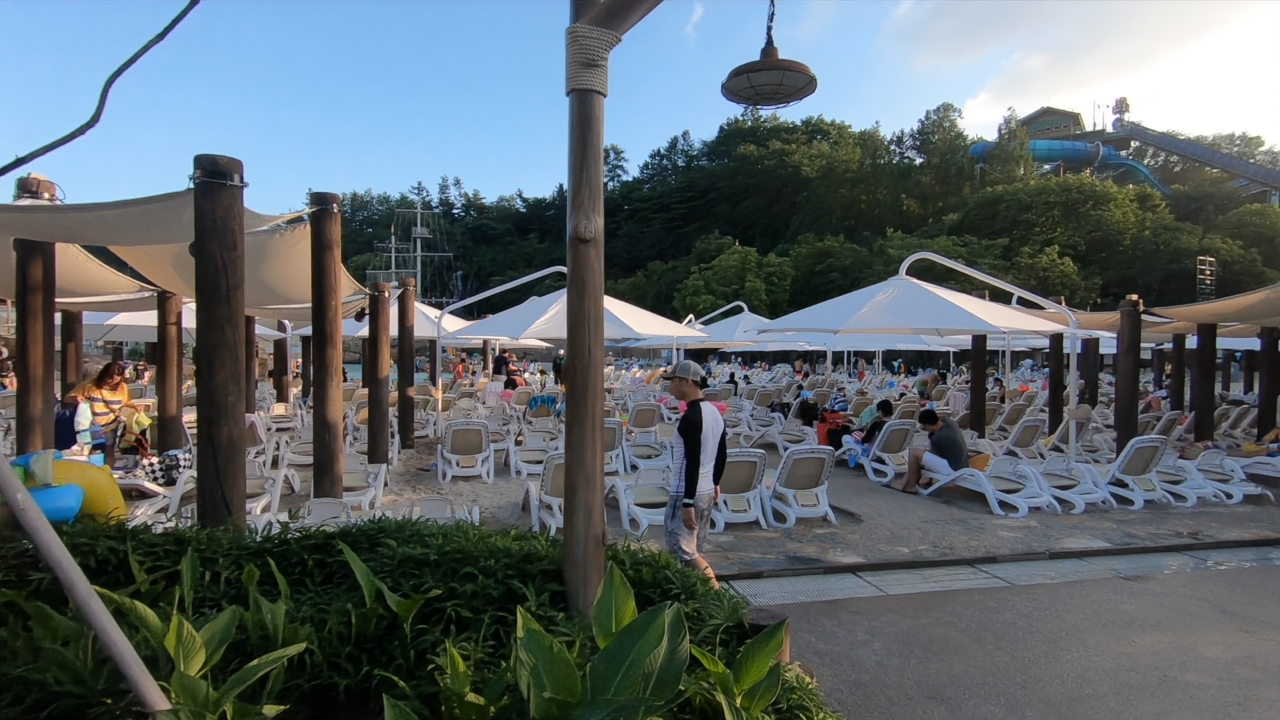}};
    		\spy on \ssxxsone in node [left] at \ssyysone;
            \spy [red] on \ssxxstwo in node [left,red] at \ssyystwo;
    	\end{tikzpicture} \hspace{-4.5mm} & 
        \begin{tikzpicture}[spy using outlines={rectangle,green,magnification=\ssmag,width=\sswspy,height=\sshspy},inner sep=0]
    		\node {\includegraphics[width=\sswidth]{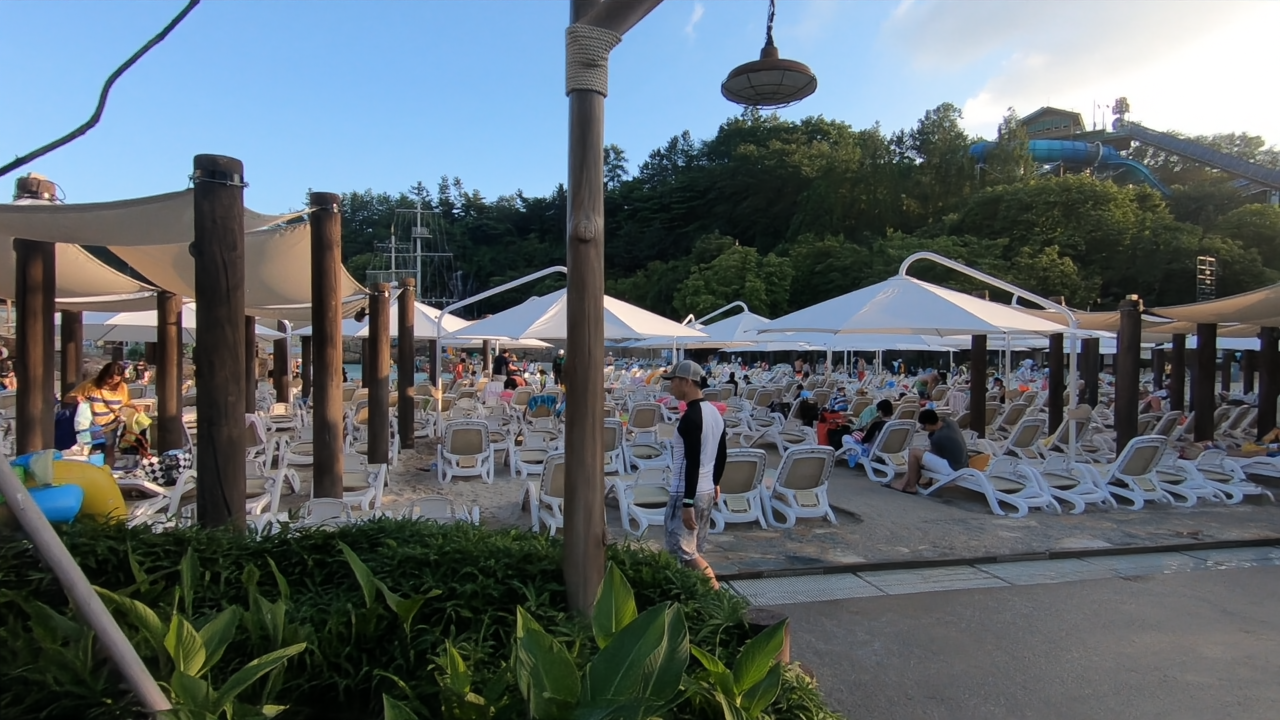}};
    		\spy on \ssxxsone in node [left] at \ssyysone;
            \spy [red] on \ssxxstwo in node [left,red] at \ssyystwo;
    	\end{tikzpicture} \hspace{-4.5mm} & 
        \begin{tikzpicture}[spy using outlines={rectangle,green,magnification=\ssmag,width=\sswspy,height=\sshspy},inner sep=0]
    		\node {\includegraphics[width=\sswidth]{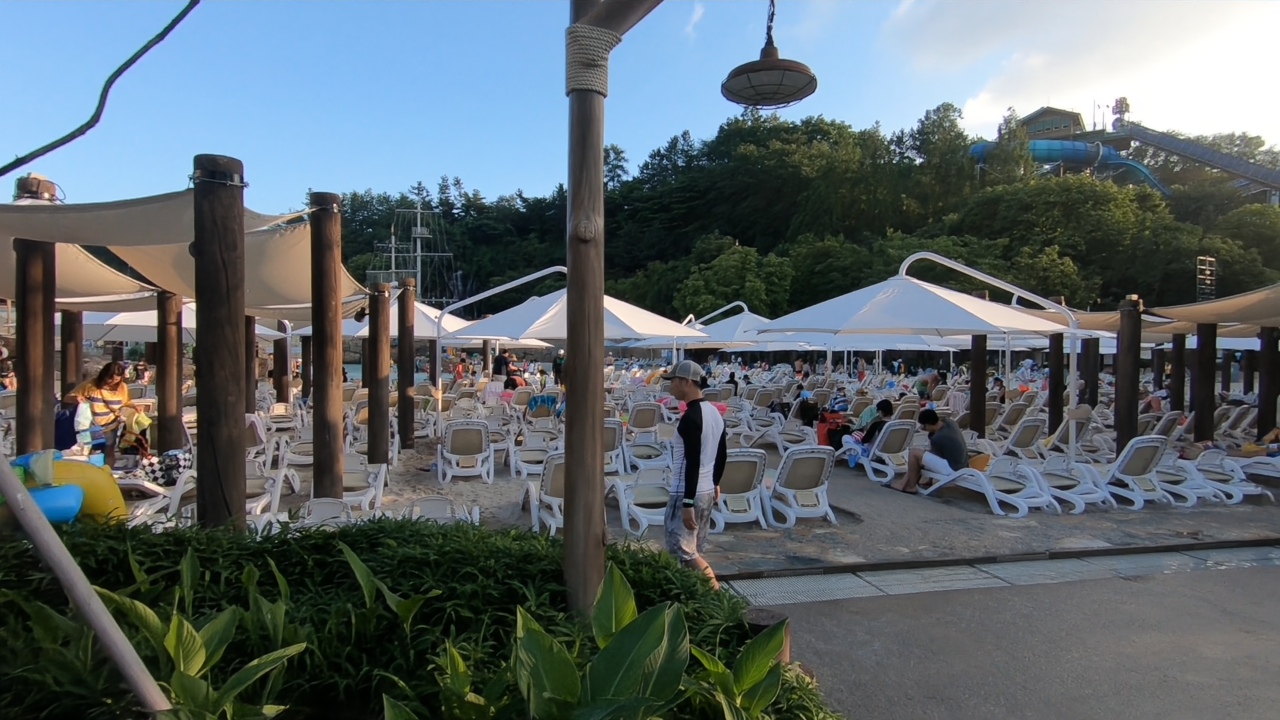}};
    		\spy on \ssxxsone in node [left] at \ssyysone;
            \spy [red] on \ssxxstwo in node [left,red] at \ssyystwo;
    	\end{tikzpicture} \vspace{-1mm} \\
        \hspace{-3mm}
        \begin{tikzpicture}[spy using outlines={rectangle,green,magnification=\ssmag,width=\sswspy,height=\sshspy},inner sep=0]
    		\node {\includegraphics[width=\sswidth]{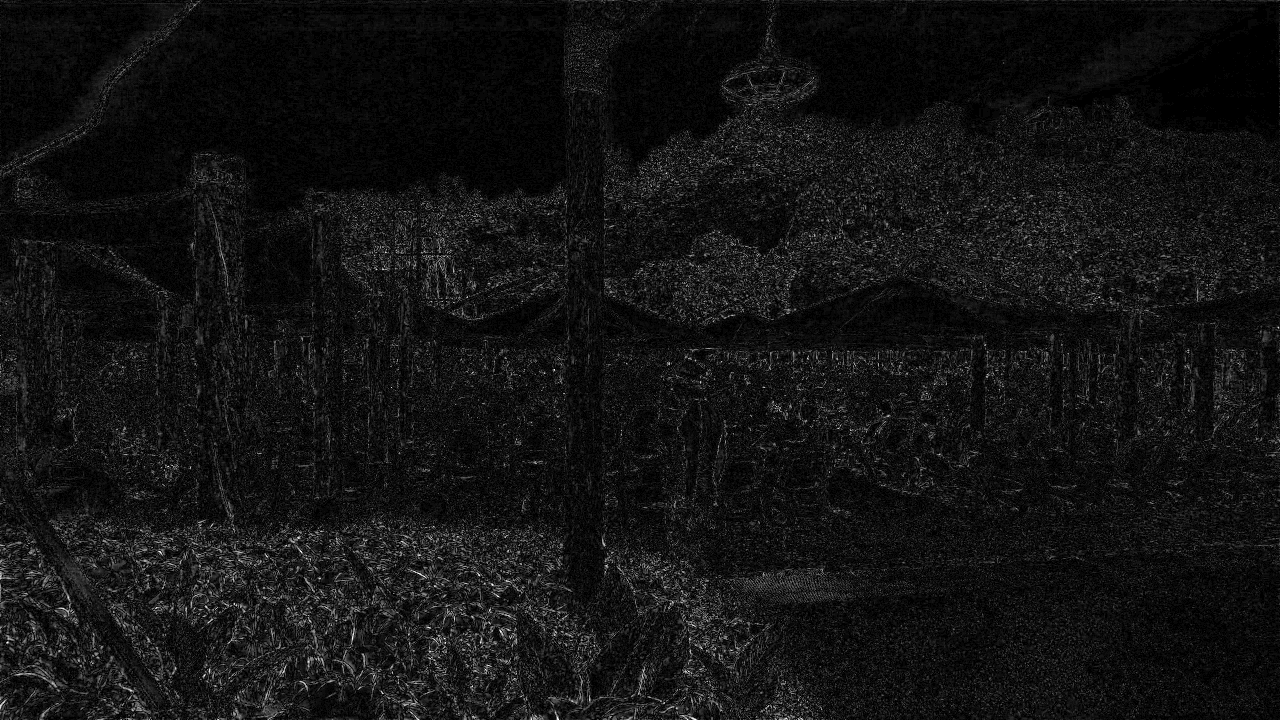}};
    		\spy on \ssxxsone in node [left] at \ssyysone;
            \spy [red] on \ssxxstwo in node [left,red] at \ssyystwo;
    	\end{tikzpicture} \hspace{-4.5mm} & 
        \begin{tikzpicture}[spy using outlines={rectangle,green,magnification=\ssmag,width=\sswspy,height=\sshspy},inner sep=0]
    		\node {\includegraphics[width=\sswidth]{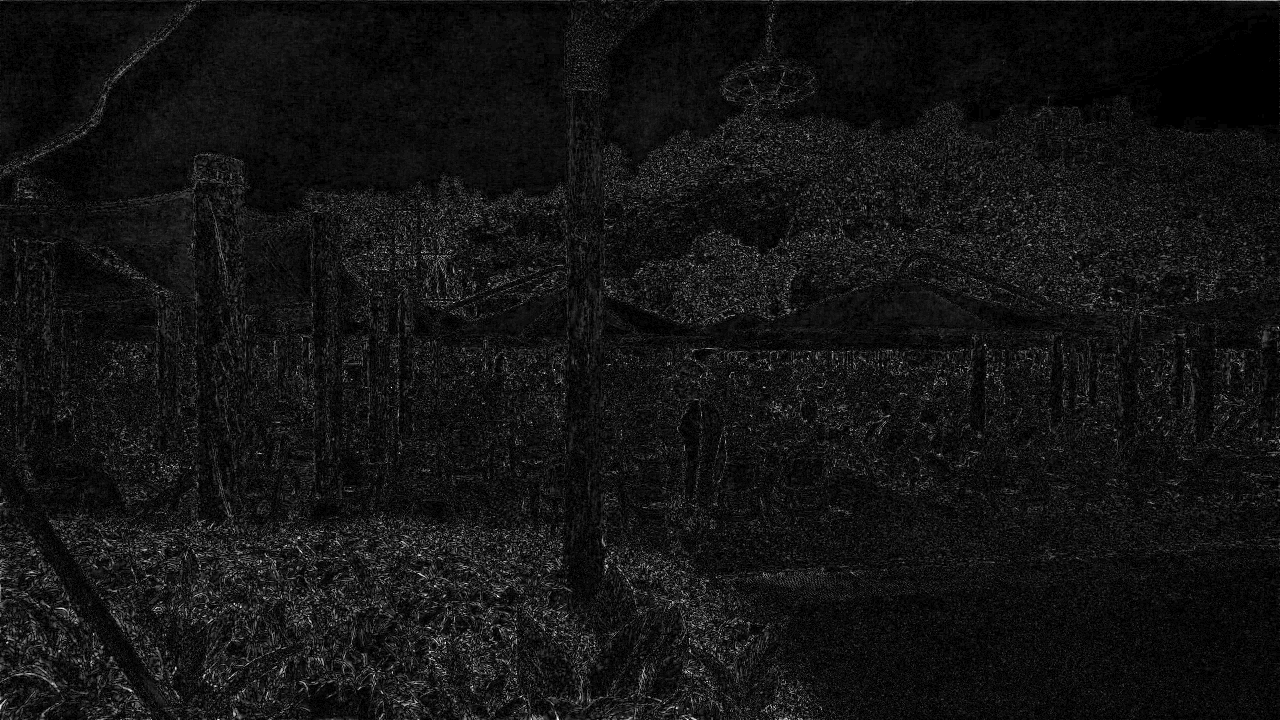}};
    		\spy on \ssxxsone in node [left] at \ssyysone;
            \spy [red] on \ssxxstwo in node [left,red] at \ssyystwo;
    	\end{tikzpicture} \hspace{-4.5mm} & 
        \begin{tikzpicture}[spy using outlines={rectangle,green,magnification=\ssmag,width=\sswspy,height=\sshspy},inner sep=0]
    		\node {\includegraphics[width=\sswidth]{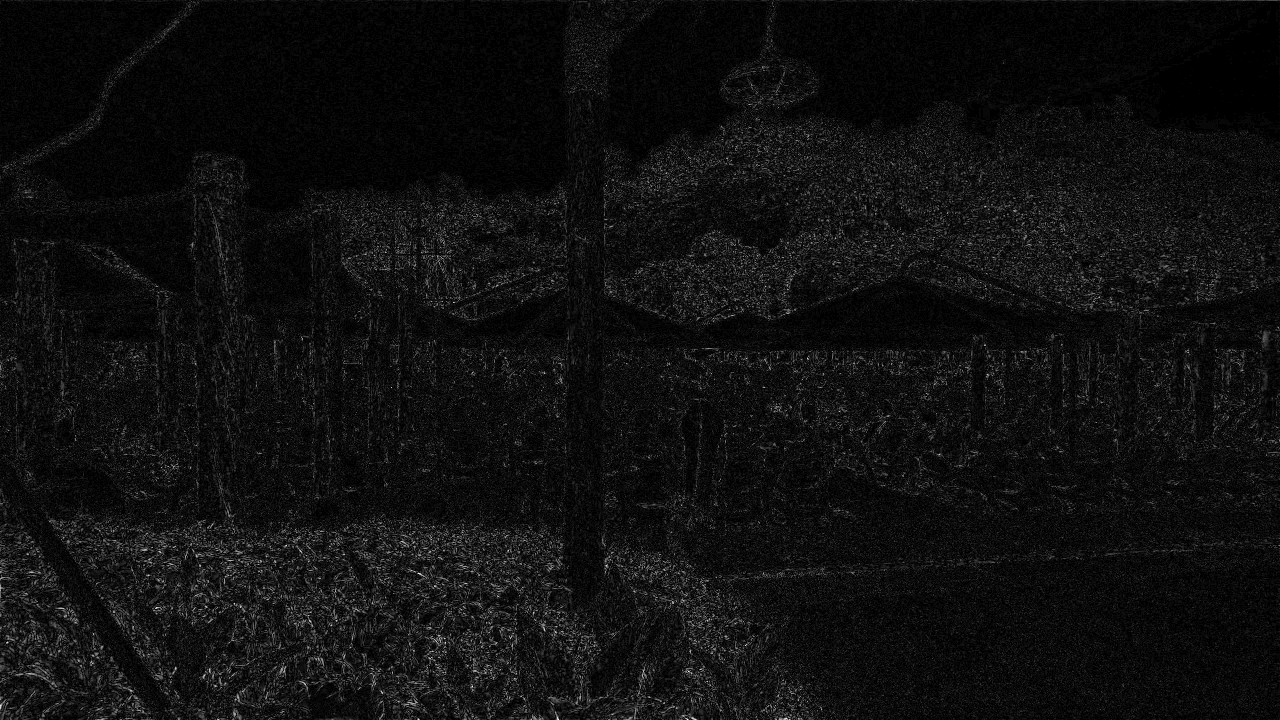}};
    		\spy on \ssxxsone in node [left] at \ssyysone;
            \spy [red] on \ssxxstwo in node [left,red] at \ssyystwo;
    	\end{tikzpicture} \hspace{-4.5mm} & 
        \begin{tikzpicture}[spy using outlines={rectangle,green,magnification=\ssmag,width=\sswspy,height=\sshspy},inner sep=0]
    		\node {\includegraphics[width=\sswidth]{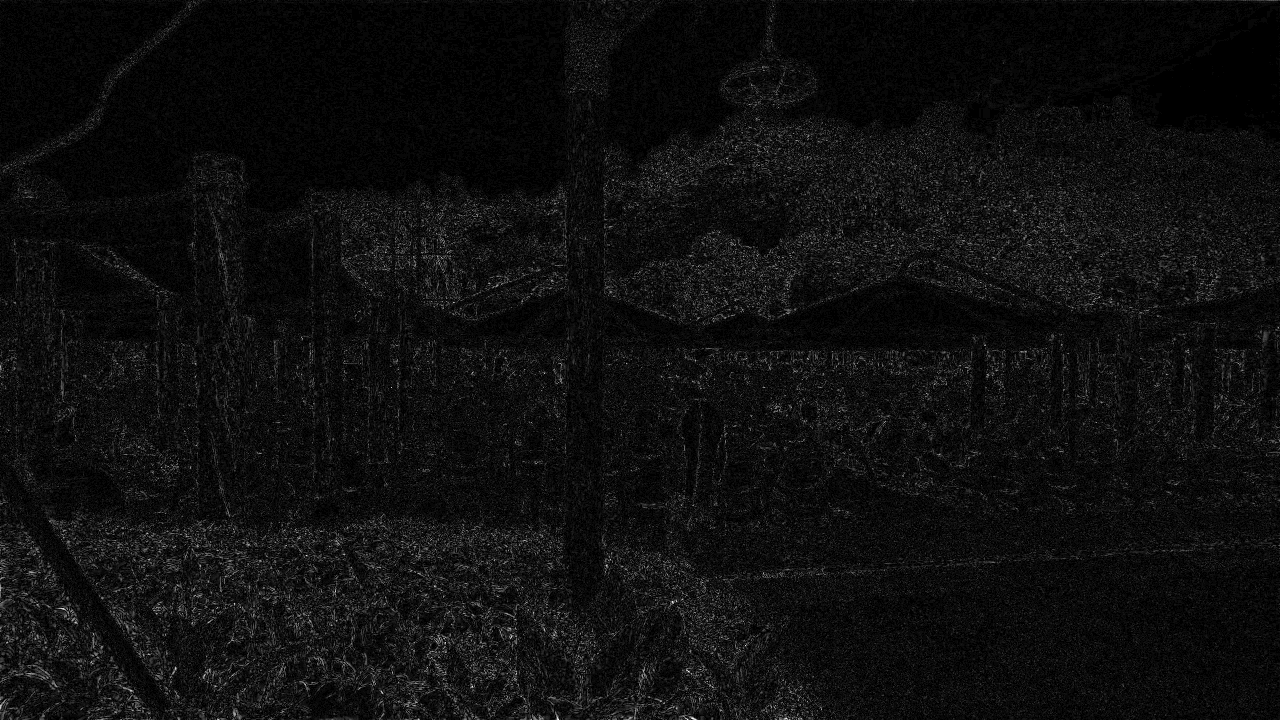}};
    		\spy on \ssxxsone in node [left] at \ssyysone;
            \spy [red] on \ssxxstwo in node [left,red] at \ssyystwo;
    	\end{tikzpicture} \vspace{-1mm} \\
        {\small w/o DA \& CAF} & {\small w/o CAF} & {\small w/o DA} & {\small w/ all} \vspace{-1mm} \\
        {\small (36.148, 0.9556)} & {\small (36.447, 0.9642)} & {\small (37.627, 0.9682)} & {\small \textbf{(38.238, 0.9730)}} \\
    \end{tabular}
    \caption{\revision{Qualitative ablations of the backbone and modules of our network and their absolute differences to the ground-truth clear image. Enc. and Dec. indicate the architecture of the encoder and decoder.}}
    \label{fig:abla_network}
\end{figure*}

\subsubsection{\revision{Effectiveness of Parallel Interactive Structure}}
\revision{As desribed in \cref{sec:task_decomposition}, an intuitive approach to solve the \myname task is to adopt two dedicated networks for the two sub-tasks, respectively, with the subsequent output fusion (detailed in the supplementary material). The major difference between such serial pipeline and our parallel design is that the former achieves the two sub-tasks independently, while the latter realize the feature interaction and enhancement progressively. Due to the degraded inputs, the serial design may introduce respective noise from each sub-task, which may further deteriorate the final result. In contrast, our parallel design can enhance the performance and alleviate the above issues. Results in \cref{tab:abla_design} show that the parallel design of \myname achieves better performance than the serial design, which verifies our inference. Moreover, the parallel design of \myname has fewer parameters than the serial design (9.46 M \vs 11.39 M), which demonstrates the efficiency of the proposed parallel architecture.}

\subsubsection{Importance of Dual-Path Backbone and DFAF Module} 
\revision{The main structure of \myname network is designed as a dual-path parallel architecture, which is composed of two parallel encoder-decoder structures, \ie, the \textit{Deblurring} and \textit{Enhancement} paths. Features of the two paths are aligned and fused gradually via the proposed Dual-path Feature Alignment and Fusion (DFAF) modules. In this section, we conduct ablation studies on the $\hat{L}_s$ to validate the effectiveness of the dual-path backbone and the DFAF module. Quantitative and qualitative results are shown in \cref{tab:abla_network,fig:abla_network}.} As shown in Exs. 1, 2, and 6, the dual-path architectures are considered as the multi-dimensional fusion for misaligned data, outperforming the single-path ones that simply concatenate the dual-exposure image and event features. 

The performance degradation from Ex. 6 to Ex. 5 validates that the Deformable Alignment (DA) block can implicitly warp the long-exposure features from the \textit{Deblurring} path $F_l$ to the short-exposure features from the \textit{Enhancement} path $F_s$ via the inter-frame motion information recorded by events. As validated by comparing Exs. 3 and 5 (or Exs. 4 and 6), the Cross-Attention Fusion (CAF) block can boost the performance since it can effectively fuse the aligned features of the \textit{Deblurring} path $F'_l$ and features of the \textit{Enhancement} path $F_s$. Without the CAF block, the absolute difference of the imaging result to the ground-truth image significantly increases, \eg, the sky color within the red box of \cref{fig:abla_network}.

\revision{The intermediate features before and after the DFAF Module are shown in \cref{fig:feature}. The qualitative comparison between $F_l$ and $F'_l$ (\cref{fig:feature}\subcref{(b)} and \subcref{(c)}) shows that the DA block provides more attention to the image textures and edgs rather than to the smooth areas. It demonstrates that with the motion information from events and the temporl reference from the short-exposure features, the DA block can implicitly realize the precise spatial alignment from the domain of the long-exposure image to that of the short-exposure one. Meanwhile, \cref{fig:feature}\subcref{(e)} and \subcref{(f)} shows that the CAF block effectively enhances the features $F_s$ with the intensity reference offered by the aligned feature $F'_l$.}

\def\sswidth{0.33\linewidth}
\begin{figure}
    \centering
    \begin{tabular}{ccc}
        \hspace{-2mm}\includegraphics[width=\sswidth]{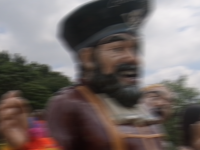} & \hspace{-3.5mm}\includegraphics[width=\sswidth]{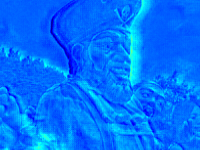} & \hspace{-3.5mm}\includegraphics[width=\sswidth]{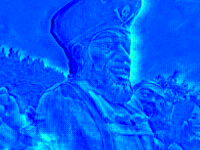} \\
        {\small (a) Long-exposure} & {\small (b) $F_l$} & {\small (c) $F'_l$} \\
        \hspace{-2mm}\includegraphics[width=\sswidth]{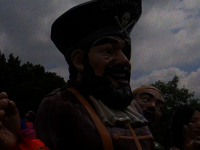} & \hspace{-3.5mm}\includegraphics[width=\sswidth]{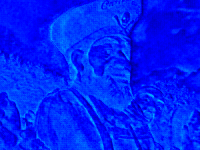} & \hspace{-3.5mm}\includegraphics[width=\sswidth]{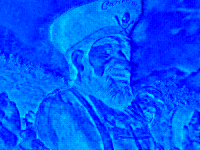} \\
        {\small (d) Short-exposure} & {\small (e) $F_s$} & {\small (f) $F'_s$} \\
    \end{tabular}
    \caption{\revision{Visualization of intermediate features from the DFAF module of our \myname. (a) and (d) the input dual-exposure image pairs. (b-c) $F_l$ and $F'_l$ are the features before and after the DA block in the \textit{Delurring} path. (e-f) $F_s$ and $F'_s$ are the features before and after the CAF block in the \textit{Enhancement} path.}}
    \label{fig:feature}
\end{figure}

\subsubsection{\revision{Impact of Balancing Parameter}}
\label{sec:lambda} 
\begin{figure}
    \centering
    \includegraphics[width=0.825\linewidth]{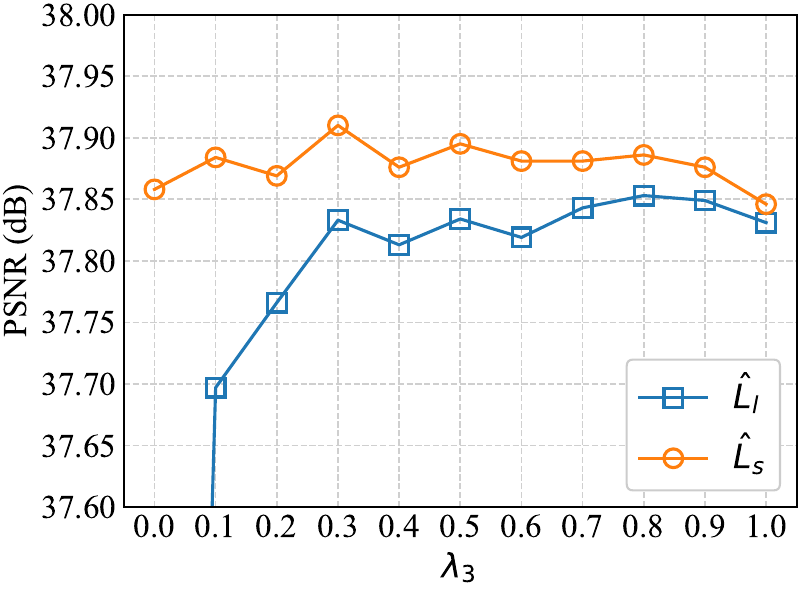}
    \caption{\revision{Evaluation of our \myname supervised by \cref{eq:loss} with the different values of the balancing parameter $\lambda_3$ in the first stage training.}}
    \label{fig:abla_lambda}
\end{figure}

\revision{Since our \myname network is supervised by the combination of the losses of the two paths in the first stage training, we validate the impact of the balancing parameters $\lambda_2$ and $\lambda_3$ on performance. We fix the $[\lambda_1,\lambda_2]=[0,1]$ and adjust various $\lambda_3$ and conduct multiple training sessions on the REDS-LL dataset. As demonstrated in \cref{fig:abla_lambda}, our \myname network obtains the optimal imaging results, \ie, 37.910 dB in the term of PSNR, when adopting $\lambda_3=0.3$. When applying $\lambda_3=1$, the result of \textit{Enhancement} path $\hat{L}_s$ outperforms that of the \textit{Deblurring} path $\hat{L}_l$, proving that the short-exposure images have more restoration potential than the long-exposure ones. Further, adopting dual-path supervision ($\lambda_3>0$) generally yields better results compared to single-path supervision ($\lambda_3=0$), which demonstrates that decomposing the complex event-based dual-exposure imaging task into the combination of two sub-tasks, \ie, event-based motion deblurring and low-light image enhancement tasks, can significantly enhance the imaging results.}

\subsection{\revision{Robustness under Various Conditions}}
\subsubsection{\revision{Robustness to Temporal Misalignment}}
\revision{The task of \myname is proposed via \cref{eq:edei_d1,eq:edei}, assuming that the cross-modal image pairs and events, are stritly aligned in the temporal domain. However, there may exist the temporal misalignment due to the hardware limitations in real-world scenes. To evaluate the robustness of our \myname network to such non-ideal conditions, we conduct a sensitivity analysis by testing the performance of \myname with the different event temporal window lengths. Specifically, we first define the exposure interval as the time duration between the short- and long-exposure images, $T_i = t_b-t_s$. In the experiments, the window length varys from $-20\%$ to $+20\%$ of the exposure interval, \ie, $\mathcal{E}_{[t_s-\varepsilon \cdot T_i, t_e]}$, where $\varepsilon\in[-20\%, +20\%]$. As shown in \cref{fig:robustness}\subcref{(a)}, the performance drop of \myname is within 1\% in PSNR, demonstrating its robustness to such misalignment. Besides, it can be observed that, in most cases, the impact of the longer event windows ($\varepsilon>0$) is less severe than that of shorter ones ($\varepsilon<0$). This is because events with a longer window still provide the reference information for the target timestamp into the network, whereas shorter windows result in the complete loss of those information.}

\subsubsection{\revision{Generalization to Unseen Exposure-Time Ratios}}
\begin{figure}
    \centering
    \begin{tabular}{c c}
        \hspace{-2mm}\includegraphics[height=3.25cm]{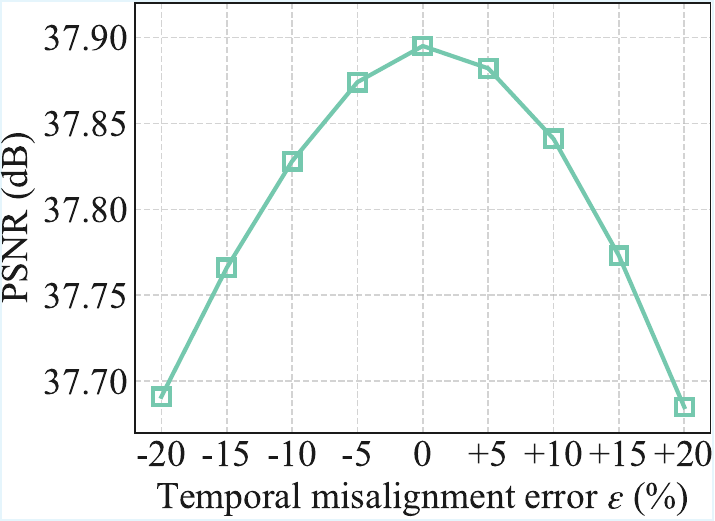}\hspace{-3.5mm} & \includegraphics[height=3.25cm]{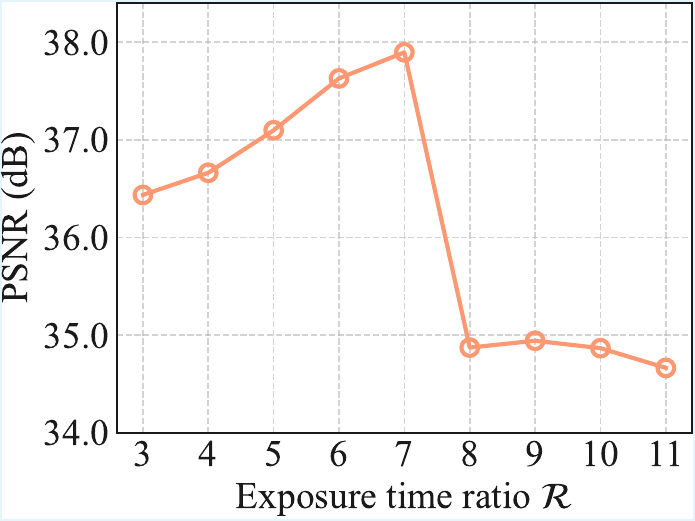} \\
        {\small (a)} & {\small (b)}
    \end{tabular}
    \caption{\revision{(a) Sensitivity analysis of the \myname network to the temporal misalignment between cross-modal image pairs and event streams. (b) Generalization of the \myname network to the unseen exposure-time ratios.}}
    \label{fig:robustness}
\end{figure}

\revision{The exposure-time ratio $\mathcal{R}$ is the ratio of the exposure time of the long-exposure image to that of the short-exposure image. In the main text, we train and test our \myname network with the same exposure-time ratio, \ie, $\mathcal{R}=7$, on the REDS-LL dataset. To evaluate the generalization of our \myname network to unseen exposure-time ratios, we test the trained model on the unseen ratios $\mathcal{R}=\{3,4,5,6,8,9,10,11\}$. Note that we maintain the time interval $T_i$ between the short- and long-exposure images unchanged. The results on the REDS-LL dataset are shown in \cref{fig:robustness}\subcref{(b)}. On the one hand, our \myname network experiences a significant drop in imaging performance when the exposure-time ratio is too large ($\mathcal{R}>7$) which may be caused by the increased difficulty in handling the severe motion blur and misalignment under such conditions. On the other hand, the imaging ability drop of our \myname network under the smaller ratios ($\mathcal{R}<7$) is relatively slight and acceptable, since the motion blur and misalignment are less severe under such conditions. We will further investigate the generalization of our \myname network to a wider range of exposure-time ratios in the future.}

\begin{table}[t]
    \centering
    \caption{\revision{Quantitative comparisons of the network parameter and inference efficiency on the REDS-LL datasets. The items of runtime and FLOPs are evaluated with the input images and events at the resolution of $1280 \times 720$.}}
    \revision{\begin{tabular}{l c c c c}
        \hline
        \multirow{2}{*}{Method} & \multirow{2}{*}{Event} & Param & Runtime & FLOPs \\
         & & (M) $\downarrow$ & ($ms$) $\downarrow$ & (G) $\downarrow$ \\
        \hline
        Uretinex-Net \cite{wu2022uretinex} & \ding{55} & \textbf{0.34} & 123.7 & 1601.25 \\
        EMNet \cite{ye2023glow} & \ding{55} & 12.52 & 774.4 & 1036.33 \\
        EvLowLight \cite{liang2023coherent} & \ding{51} & 15.03 & 766.9 & 4091.00 \\
        EvLight \cite{liang2024towards} & \ding{51} & 22.73 & 674.6 & 2731.53 \\
        \hline
        MSSNet \cite{kim2022mssnet} & \ding{55} & 15.59 & \textbf{5.4} & 2040.46 \\
        EFNet \cite{sun2022event} & \ding{51} & 8.47 & 30.4 & 1712.25 \\
        STCNet \cite{yang2024motion} & \ding{51} & 16.25 & 79.0 & 2611.93 \\
        \hline
        LSD$_2$ \cite{mustaniemi2020lsd2} & \ding{55} & 31.03 & 14.0 & 679.06 \\
        LSFNet \cite{chang2021low} & \ding{55} & 10.20 & 23.0 & \textbf{498.36} \\
        D2HNet \cite{zhao2022d2hnet} & \ding{55} & 79.58 & 83.7 & 1208.24 \\
        \myname (w/o CGF) & \ding{51} & 8.58 & 28.6 & 1344.76 \\
        \myname (w/ CGF) & \ding{51} & 9.46 & 33.0 & 2175.54 \\
        \hline
    \end{tabular}}
    \label{tab:param_runtime_flops}
\end{table}

\subsection{\revision{Parameter and Efficiency Analysis}}
\revision{We further evaluate the complexity of our proposed algorithm and other methods by feeding dual-exposure image pairs and event streams with the same spatial resolution $1280\times720$ and executing them on the same NVIDIA GeForce RTX 3090 GPU. The comparisons on the number of network parameters, runtime, and FLOPs are shown in \cref{tab:param_runtime_flops}. Utilizing the motion information recorded by events, our \myname performs better than the state-of-the-art F-DEI method, \ie, D2HNet \cite{zhao2022d2hnet}, with much smaller model size (9.46 M \vs 79.58 M) and less inference time (33.0 $ms$ \vs 83.7 $ms$). Our base model (w/o CGF) requires only 8.58M parameters and 1344.76 GFLOPs, operating at 28.6 $ms$ per frame. The integration of the CGF module introduces only a marginal computational overhead, increasing the parameter count to 9.46M and runtime to 33.0 $ms$, while remaining significantly lightweight and efficient. Our \myname network can infer high-quality imaging with the shorter time consumption than the majority of recent event-based methods, validating its superiority in efficiency and effectiveness.}

\section{\revision{Limitations and Future Work}}
\label{sec:limitation}
\begin{figure}[t]
    \centering
    \begin{tabular}{c c c}
        \hspace{-2mm}\includegraphics[width=0.33\linewidth]{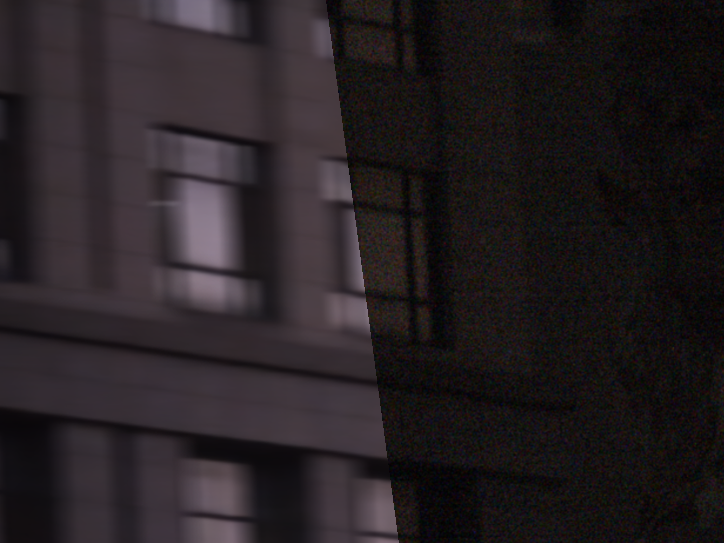}\hspace{-4mm} & \includegraphics[width=0.33\linewidth]{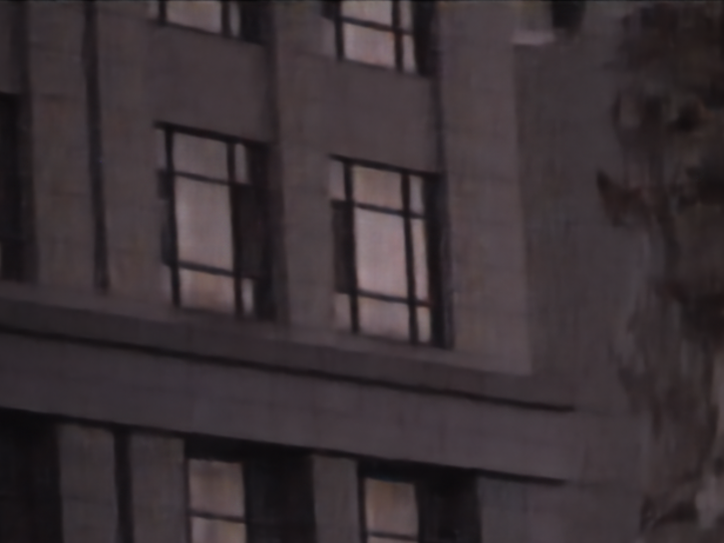}\hspace{-4mm} & \includegraphics[width=0.33\linewidth]{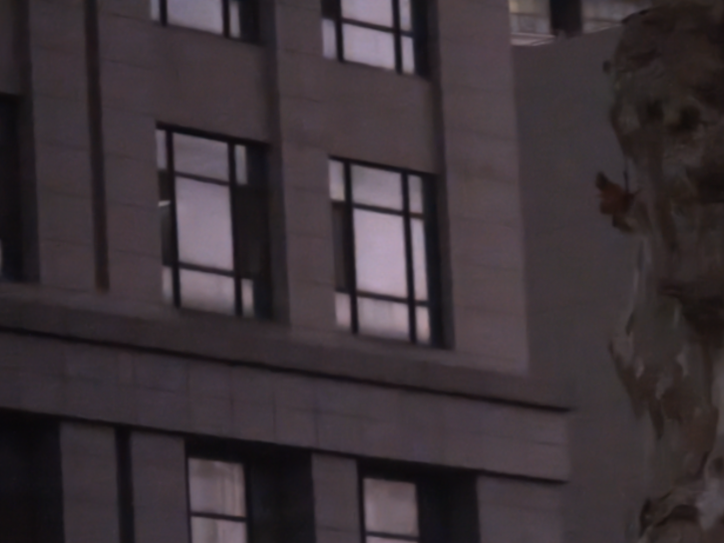}\vspace{-1mm} \\
        \hspace{-2mm}\includegraphics[width=0.33\linewidth]{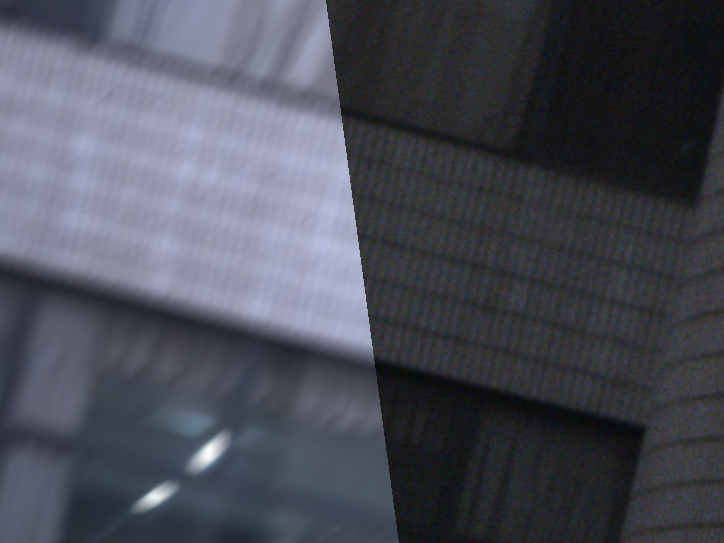}\hspace{-4mm} & \includegraphics[width=0.33\linewidth]{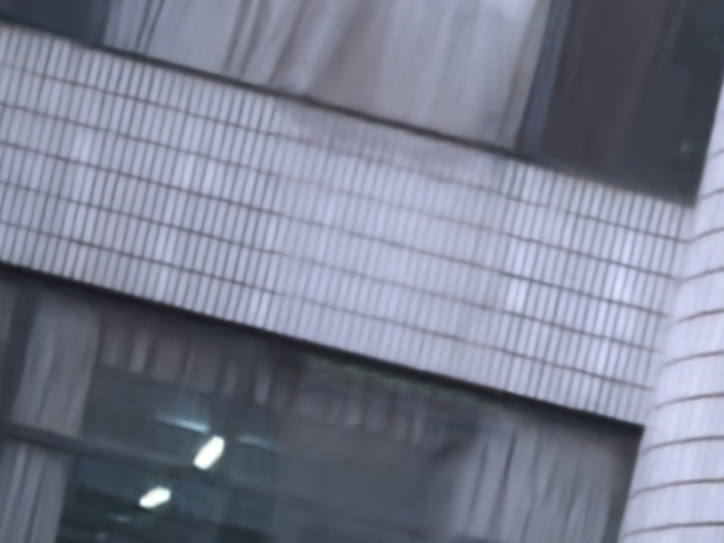}\hspace{-4mm} & 
        \includegraphics[width=0.33\linewidth]{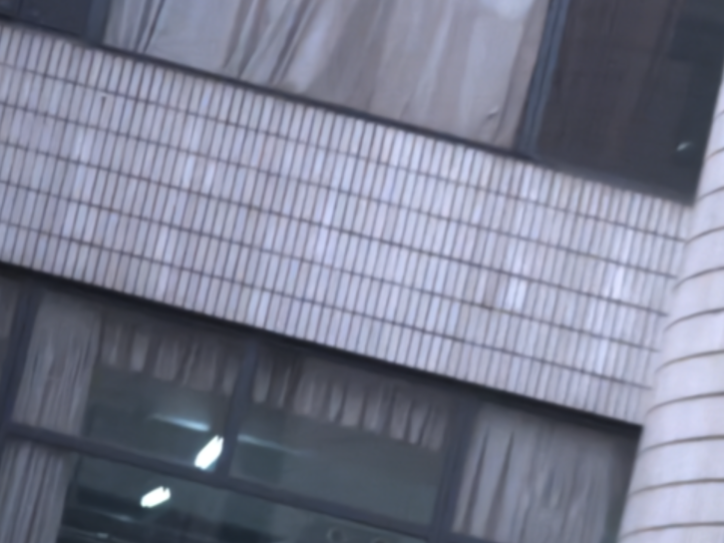}\vspace{-1mm} \\
        \hspace{-2mm}\includegraphics[width=0.33\linewidth]{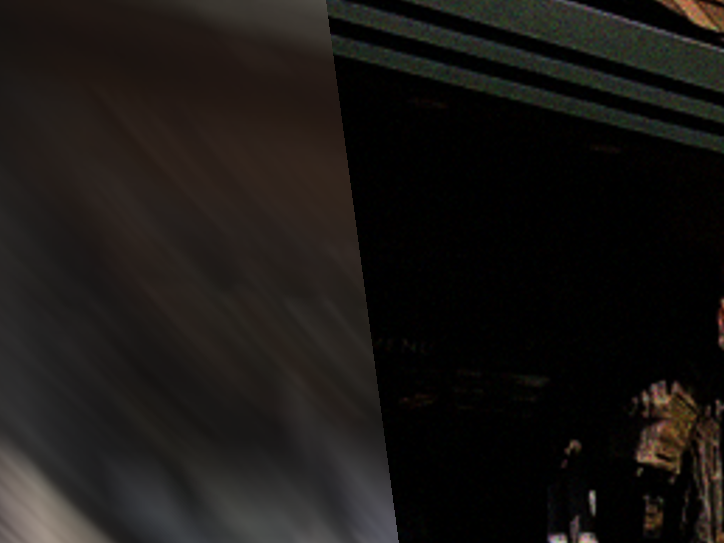}\hspace{-4mm} & \includegraphics[width=0.33\linewidth]{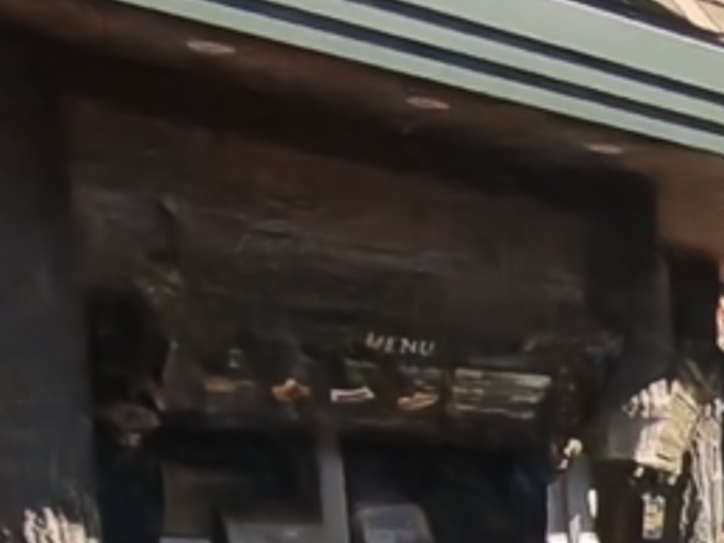}\hspace{-4mm} & \includegraphics[width=0.33\linewidth]{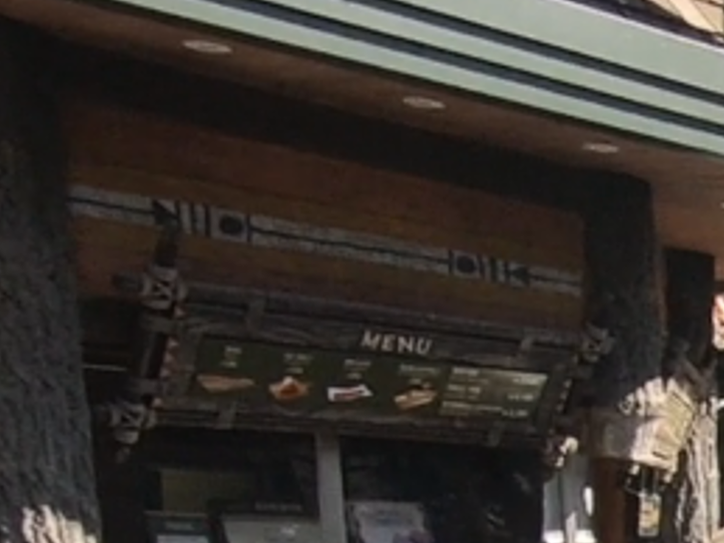} \\
        {\small (a) Image inputs} & {\small (b) \myname} & {\small (c) GT}
    \end{tabular}
    \caption{\revision{Failure cases under non-ideal scenes, \eg, extremely dim environment (upper), blocked and discarded events (middle), and coupled degradations (lower).}}
    \label{fig:fail_case_r1}
\end{figure}

\revision{As shown in \cref{fig:fail_case_r1}, the proposed \myname may suffer from performance degradation under non-ideal scenes: \textbf{1) Extremely dim environment.} The sensor of event cameras has a lower cutoff frequency, leading to the trailing event effect and the loss of intricate texture details \cite{hu2021v2e}. As shown in the first row, it causes inaccurate alignment of the double-exposure image pairs, resulting in severe color aliasing distortion. \textbf{2) Impact of blocked and discarded events.} Fast motion across dense textures may cause the event camera to generate a large number of events, which may exceed the bandwidth of the data transmission and lead to blocked and discarded events. As shown in the second row, our \myname struggles to perform precise spatial alignment in the event-depleted areas, leading to severe artifacts. \textbf{3) Coupled degradations.} The third row shows the impact of coupling interference caused by the severe blur due to high-speed motion and the dark region caused by rooftop shadow. The proposed \myname fails to recover reliable texture details. We plan to address these issues in the future.}

% \begin{figure}[t]
%     \centering
%     \begin{tabular}{c c c}
%         \hspace{-2mm}\includegraphics[width=0.33\linewidth]{figs/failcase/RLIED_2025-03-24_18_43_10_input.png}\hspace{-3.5mm} & \includegraphics[width=0.33\linewidth]{figs/failcase/RLIED_2025-03-24_18_43_10_pred_s.png}\hspace{-3.5mm} & \includegraphics[width=0.33\linewidth]{figs/failcase/RLIED_2025-03-24_18_43_10_gt.png}\vspace{-1mm} \\
%         \hspace{-2mm}\includegraphics[width=0.33\linewidth]{figs/failcase/RLIED_2025-03-25_18_23_18_input.png}\hspace{-3.5mm} & \includegraphics[width=0.33\linewidth]{figs/failcase/RLIED_2025-03-25_18_23_18_pred_s.png}\hspace{-3.5mm} & 
%         \includegraphics[width=0.33\linewidth]{figs/failcase/RLIED_2025-03-25_18_23_18_gt.png}\vspace{-1mm} \\
%         \hspace{-2mm}\includegraphics[width=0.33\linewidth]{figs/failcase/REDS_022_input.png}\hspace{-3.5mm} & \includegraphics[width=0.33\linewidth]{figs/failcase/REDS_022_pred_s.png}\hspace{-3.5mm} & \includegraphics[width=0.33\linewidth]{figs/failcase/REDS_022_gt.png} \\
%         \hspace{-2mm}{\small (a) Image inputs} & {\small (b) \myname} & {\small (c) GT}
%     \end{tabular}
%     \caption{\revision{Failure cases under non-ideal scenes, \eg, extremely dim environment (upper), over-exposure input (middle), and coupled degradations (lower).}}
%     \label{fig:fail_case_r1}
% \end{figure}

\section{Conclusion}
In this paper, we propose a novel \myname network to reconstruct high-quality images from the short- and long-exposure image pairs and corresponding event streams under the low-light scenarios. Specifically, we propose to decompose the complex event-based dual-exposure imaging task into a combination of two sub-tasks, \ie, the event-based motion deblurring task and low-light image enhancement task, which guides us to design our \myname as a dual-path feature propagation architecture. A core DFAF module is proposed to gradually align and fuse features from short- and long-exposure images. Furthermore, we build a hybrid camera system to collect a new dataset containing paired low- and normal-light image pairs and real-world events with diverse motion types. Extensive experiments demonstrate that our proposed \myname achieves the state-of-the-art low-light imaging performance.

% Assuming that events are triggered solely by motion, our proposed \myname may be subject to events from other sources, \eg, illumination change and light flickering \cite{wang2022linear}. Additionally, in extremely dim environments (\eg, $<1$ lux), the vision sensor of the event camera has a lower cutoff frequency, leading to the trailing event effect and the loss of intricate texture details \cite{hu2021v2e}. It may mislead our algorithm in aligning long- and short-exposure images. We plan to address these issues in the future.

\bibliographystyle{IEEEtran}
\bibliography{main.bib}

\end{document}